\documentclass{article}

\usepackage{iclr2025_conference_arxiv,times}

\usepackage[utf8]{inputenc} 
\usepackage[T1]{fontenc}    
\usepackage[bookmarks=false,hidelinks]{hyperref}       
\usepackage{url}            
\usepackage{booktabs}       
\usepackage{amsfonts}       
\usepackage{nicefrac}       
\usepackage{microtype}      
\usepackage{xcolor}         
\usepackage{natbib}
\usepackage{microtype}
\usepackage{svg}
\usepackage{graphicx}
\usepackage{subfigure}
\usepackage{booktabs} 
\usepackage{makecell}
\usepackage{multirow}
\usepackage{enumitem}
\usepackage{wrapfig}
\usepackage{caption}
\setcitestyle{authoryear,open={},close={},citesep={;},aysep={,},yysep={;}}

\usepackage{algorithm}
\usepackage{algorithmicx}
\newcommand{\longscore}{\underline{\hspace{0.18cm}}}

\usepackage{amsmath}
\usepackage{amssymb}
\usepackage{mathtools}
\usepackage{amsthm}
\usepackage[capitalize,noabbrev]{cleveref}
\usepackage{accents}

\newcommand{\method}{\text{Activation Addition}}
\newcommand{\shortmethod}{\text{ActAdd}}

\newcolumntype{P}[1]{>{\centering\arraybackslash}p{#1}}
\newcommand\thickbar[1]{\accentset{\rule{.4em}{.8pt}}{#1}}

\definecolor{airforceblue}{rgb}{0.36, 0.54, 0.66}

\theoremstyle{plain}

\theoremstyle{definition}

\theoremstyle{remark}

\usepackage[textsize=tiny]{todonotes}
\newcommand{\oblong}[1][]{
\begin{tikzpicture}[#1]
    \draw[rounded corners] (0, 0) rectangle (0.5, 0.3) {};
\end{tikzpicture}
}
\definecolor{activationcolor}{HTML}{1BA1E2}

\title{Steering Language Models With Activation Engineering}

\author{%
  Alexander Matt Turner \\
  Independent researcher\\
  \texttt{alex@turntrout.com} \\
  \And
  Lisa Thiergart \\
  MIRI \\
  \And
  Gavin Leech \\
  University of Bristol \\
  \texttt{g.leech@bristol.ac.uk} \\
  \AND
  David Udell\\
  Independent researcher \\
  \And
  Juan J. Vazquez \\
  Arb Research \\
  \And
  Ulisse Mini\\
  MATS \\
  \And
  Monte MacDiarmid \\
  Anthropic \\
}

\iclrfinalcopy

\begin{document}

\maketitle

\begin{abstract}
Prompt engineering and finetuning aim to maximize language model performance on a given metric (like toxicity reduction). However, these methods do not fully elicit a model’s capabilities.  To reduce this gap, we introduce \emph{activation engineering}: the inference-time modification of activations in order to control (or \emph{steer}) model outputs. Specifically, we introduce the \emph{\method} (\shortmethod) technique, which contrasts the intermediate activations on prompt pairs (such as “Love” versus “Hate”) to compute a \emph{steering vector} (\citep{subramani-2022-extracting}). By tactically adding in e.g. the “Love” $-$ “Hate” steering vector during the forward pass, we achieve SOTA on negative-to-positive sentiment shift and detoxification using models including LLaMA-3 and OPT. {\shortmethod} yields inference-time control over high-level output properties (like topic and sentiment) while preserving performance on off-target tasks. {\shortmethod} is lightweight: it does not require any machine optimization and works with a single pair of data points, which enables rapid iteration over steering. {\shortmethod} demonstrates the power of activation engineering.
\end{abstract}

\section{Introduction}\label{sec:intro}

LLMs contain hidden capabilities we do not know how to fully elicit (\citep{korinek2023language}). Naively prompting a model with a question does not maximize the probability of the correct response.  Carefully consider how prompting a model to think ``step-by-step'' (\citep{wei2022chain}) massively improves performance on a range of benchmarks. Similarly, ``few-shot'' prompting a model with correct answers to unrelated in-distribution questions allows ``in-context learning'' for e.g. stronger performance on NLP tasks (\citep{brown2020language}). Importantly, these interventions do not supply the LLM with extra task-relevant information or update the algorithm implemented by the LLM's computational graph. Even though the model is initially \emph{able} to score higher on these benchmarks, those capabilities do not emerge without a specific intervention. We therefore hypothesize the presence of an \emph{elicitation overhang}: we do not know how to elicit all relevant abilities and information from frontier models.

Prompt engineering is the most obvious way to steer a model, but prompting has limited reliability (\citep{NEURIPS2022_c4025018,wang2024prompt}). Therefore, to reduce the elicitation overhang, we explore a new modality for steering language model outputs. By strategically perturbing activations during the forward pass, we hope to more reliably and effectively steer models compared to prompt engineering. We call this methodology \emph{activation engineering}.

We suspect that compared to prompt engineering, activation engineering can elicit a wider range of model capabilities. Consider, for example, a model optimized to imitate the text outputs of eloquent poets and awkward mathematicians. The model may contain the internal mechanisms required to output text which is \emph{both} eloquent and mathematical. However, if the model is an accurate estimator of the training distribution, it will (correctly) assign low probability to eloquent mathematical prose. Because nothing in the training data was both eloquent and mathematical, there may exist no prompt which elicits mathematical prose. In contrast, activation engineering might be able to simultaneously activate the circuitry for eloquent speech and for mathematical content.

To demonstrate the power of activation engineering, we introduce \emph{Activation Addition} (\shortmethod). Suppose we want to achieve negative-to-positive sentiment control (\citep{li2018deleteretrievegeneratesimple,dathathri2020plug}). To achieve this, {\shortmethod} first compares the model's activations on a contrast pair of prompts, such as the prompts ``Love'' and ``Hate.'' The otherwise-similar prompts differ along the target dimension of sentiment. {\shortmethod} then computes the difference of these activations in order to compute \emph{steering vectors}. These vectors act like ``virtual bias terms'' because {\shortmethod} \emph{directly adds} the steering vectors to the forward pass at inference time. By shifting the inference-time activations along the direction of the steering vector, {\shortmethod} steers the model to generate positive sentiment completions (\cref{tab:hate}).

\bgroup
\def\arraystretch{1}
\begin{table}[ht]
\centering
\caption{The impact of \shortmethod. {The steering vectors are computed from (``Love'' - ``Hate'') and (``I talk about weddings constantly'' - ``I do not talk about weddings constantly''). Appendix \Cref{tab:longer_examples} shows more examples.
}
}
\vspace{2mm}
\begin{tabular}{p{3cm}p{0.18cm}P{1.6cm}l}
\toprule
\thead{\textbf{Prompt}} 
&\thead{+}
& \thead{\textbf{steering}}
& \thead{\,= \qquad\qquad\qquad \textbf{completion}}
\\\hline
& \\
\begin{tabular}[p]{@{}c@{}}
    \vspace{4mm}\\
    I hate you because...
\end{tabular} 
&&
\, {[None]}
&
\begin{tabular}[p]{@{}c@{}}
    \qquad \small ...you are the most disgusting thing I have ever seen.
\end{tabular}

\\
\\

&& 
\begin{tabular}[p]{@{}c@{}}
    \shortmethod \\ (love)
\end{tabular}
&
\begin{tabular}[p]{@{}c@{}}
    \qquad \small ...you are so beautiful and I want to be with you forever. 
\end{tabular}
\\\\\hline
\multirow{ 2}{*}{
\begin{tabular}[p]{@{}c@{}}
    \vspace{3mm}\\
    I went up to my \\
    friend and said... \\
\end{tabular} }

&& 
\vspace{0.5mm}
\,\,\,[None]
& 
\begin{tabular}[p]{@{}c@{}}
    \\
    \small \qquad ...``I'm sorry, I can't help you.'' \\
    \qquad \small ``No,'' he said. ``You're not.'' 
\end{tabular}
\\\\
&&
\begin{tabular}[p]{@{}c@{}}
    \shortmethod\\ (weddings)
\end{tabular}
& 
\begin{tabular}[p]{@{}c@{}}
    \qquad\small ...``I'm going to talk about the 
wedding in this episode of \\  
\qquad\small Wedding Season. I think it's a really good episode. \\ 
\qquad\small It's   
about how you're supposed  
to talk about weddings.''
\end{tabular}
\\
\\
\bottomrule
\end{tabular}
\label{tab:hate}
\end{table}
\vspace{2mm}

\paragraph{Contributions.} We unify past literature on related topics to introduce \emph{activation engineering}. To better elicit the full capabilities of models, we introduce the {\shortmethod} steering method, which achieves SOTA on toxicity reduction and sentiment control. We thoroughly test the steered models to verify the preservation of their general capabilities. We therefore show the promise of {\shortmethod} as an effective and cheap method for steering LLM outputs.

\section{Related Work}

\textbf{Latent space arithmetic.}
Research in generative models for computer vision has long demonstrated the ability to steer image generation using derived vectors, including steering latent variables -- most famously, shifting activations along a direction that corresponds to smiles in images (\cite{larsen2016autoencoding,white2016sampling}). Similarly, in the text domain, classic results on the word2vec embedding show that arithmetic on word vectors can capture some parts of semantic reasoning (for instance, analogies: \cite{mikolov2013linguistic,NIPS2013_9aa42b31}). Our work focuses on steering generative language models.

\paragraph{LLM steering.} Many approaches attempt to affect the output of a pretrained LLM, whether:
\begin{itemize}[leftmargin=*]
    \item \textit{Intervening on weights}, as with supervised finetuning, RLHF, steerable layers, and weight editing (that is, targeted fine-tuning) (\cite{ranzato2016sequence,ziegler2020finetuning,dathathri2020plug,meng2023locating, ilharco2023editing}).
However, naive RLHF, finetuning, and weight editing have known side-effects on overall model performance (\cite{hase2023does,qi2023fine, brown2023robustness});
    \item \textit{Intervening at decoding}, as with guided or trainable decoding (\cite{gu-etal-2017-trainable,grover2019bias}; see \citealt{zhang2022survey} for an overview of controlled generation and \citealt{jin-etal-2022-deep} for textual style transfer);
    \item \textit{Intervening on the prompt}, as with automated prompt engineering (\cite{shin-etal-2020-autoprompt,zhou2022steering});
    \item \textit{Intervening on token embeddings}, as with `soft prompting' (\cite{li2021prefixtuning,lester2021power,khashabi-etal-2022-prompt});
    \item \textit{Intervening on activations}, for instance by freezing the weights of the LLM and searching for a `steering vector' of activations, e.g. using gradient descent (\cite{subramani-2022-extracting,hernandez2023inspecting}). These optimized extraction methods, which search for a steering vector, differ from 
    extraction methods which directly compute it (present work and \citealt{li2023inferencetime}). In our work, we do not use gradient descent or other optimization methods.
\end{itemize}

\bgroup
\def\arraystretch{1.5}
\begin{table}[ht]
\caption{Locating our work in the steering literature.}
\centering

\begin{tabular}{c|cc}
\toprule
     & \multicolumn{2}{c}{\textbf{\begin{tabular}[c]{@{}c@{}}Vector intervenes on model ...\end{tabular}}}
     \\
\textbf{Intervention vectors obtained via}

& \multicolumn{1}{c}{... \textit{weights}}             & \multicolumn{1}{c}{... \textit{activations}}  \\\hline
\begin{tabular}{@{}c@{}}
Differences after fine-tuning
\end{tabular}
& 
Ilharco 2023
& 

N/A  
\\\hline
\begin{tabular}{@{}c@{}}
Per-query gradient-based search
\end{tabular}
&               
\begin{tabular}{@{}c@{}}
Meng 2022,\\ 
Orgad 2023
\end{tabular}  
& 
\begin{tabular}{@{}c@{}}
Dathathri 2020\\
Subramani 2022\\
Hernandez 2023
\end{tabular}

\\\hline
\begin{tabular}{@{}c@{}}Differences between prompt pairs                 
\end{tabular}
&                                      
N/A      
& 

 \begin{tabular}{@{}c@{}}
\textbf{\shortmethod\,} (present work),\\ 
\citep{li2023inferencetime} 
\end{tabular}
\\

\bottomrule
\end{tabular}
\label{tab:steering_lit}
\end{table}
\egroup


\paragraph{Activation engineering.}
Activation engineering involves creating vectors of activations which cause desired changes to output text when added to the forward passes of a frozen LLM (\cite{dathathri2020plug}). \Cref{tab:steering_lit} organizes prior work by intervention type. 

An early antecedent is the Plug-and-Play Language Model of \citealt{dathathri2020plug}. This uses a separate classifier (one classifier per attribute to steer towards) to perturb the
model's activations to generate text that accords more closely with the classifier's target.
\citealt{subramani-2022-extracting} extract latent steering vectors from a frozen LLM, successfully discovering sentence-specific vectors which steer completions to near-perfect BLEU scores (i.e, control of the LLM's generation) and unsupervised style transfer. However, the method requires running gradient descent for each new steering vector.
\citealt{hernandez2023inspecting} locate and edit an LLM's knowledge through learning an encoding of facts in its activation space. Ablating attention heads can also be seen as activation engineering, though the technique is mostly used for model interpretation rather than steering (\cite{NEURIPS2019_2c601ad9,olsson2022context}).

Independently, \citealt{li2023inferencetime} developed a similar method called ITI which computes steering vectors which are selectively applied according to trained linear probes. They use these probes to find attention heads with different activation distributions for true and false statements. They steer the model toward truthful outputs, where our experiments cover a range of goals. In addition, ITI adds the same steering vector at all sequence positions during inference and ITI requires dozens of samples. In contrast, {\shortmethod} we add steering vectors to a subset of sequence positions and require as few as 2 samples. Similar work on `in-context vectors' also followed ours (\cite{liu2023incontext}). Lastly, \citet{zou2023representation}'s ``representation engineering'' also followed our work. They develop a range of techniques for deriving steering vectors and for steering models using activation-space edits and optimization. In comparison to \citet{zou2023representation}, we steer different models (LLaMA-3, OPT, GPT-2, and GPT-J) on different tasks (detoxification and sentiment control).

\begin{figure*}
    \centering
    \includegraphics[width=\textwidth]{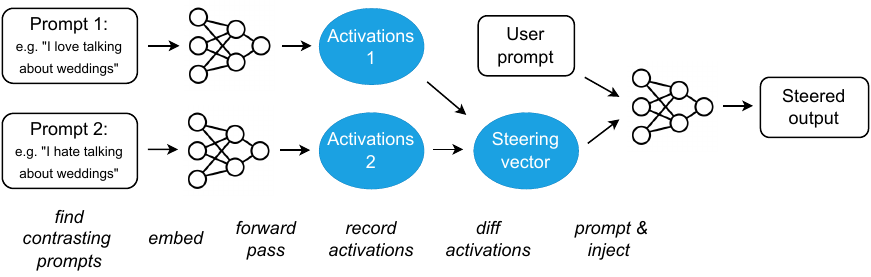}
    
    \caption{Schematic of the \method\,(\textbf{\shortmethod}) method. 
    \oblong[] $=$ natural language text; 
    {\huge\color{activationcolor}$\bullet$\,} = vectors of activations just before a specified layer. In this example, the output is heavily biased towards discussing weddings, regardless of the topic of the user prompt. (See Algorithm~\ref{alg} for the method's parameters: intervention strength,  intervention layer, and sequence alignment.)}
    \label{fig:method}
\end{figure*}

\section{How Activation Addition works}
\label{sec:meth}

We use decoder-only Transformer neural networks (\cite{NIPS2017_3f5ee243}). The LLMs in this work contain a stack of Transformer layers, each consisting of multi-head attention (MHA) and a feedforward network (FFN). We focus on its ``residual streams'' (\cite{elhage2021mathematical}),  the sequences $(\mathbf{x}_0, ... , \mathbf{x}_n)$ of intermediate activation vectors processed by each layer. {\shortmethod} manipulates the residual stream values $\mathbf{h}^l$ input to layer $l$. Each layer performs MHA and FFN computations on $\mathbf{x}_i$, adding $\mathbf{x}_{i+1}$ to the stream. The final vector $\mathbf{x}_n$ in the stream can then be decoded into the next-token prediction. At inference time, the residual stream is initialized $\mathbf{h}^1$ with the embedding of the tokenized prompt. 

\textbf{Activation addition.}\quad
Our method takes a pair of natural-language prompts $(p_+, p_-)$, where $p_+$ represents the property we wish output text to emphasise (e.g. love) and $p_-$ represents its opposite (e.g. hate or indifference). $\mathbf{h}_+^l$ is the activation vector for the prompt $p_+$ at layer $l$. The difference $\mathbf{h}_+^l - \mathbf{h}_-^l $ is a new activation vector which (intuitively) captures the difference between a prompt with the target property, and a prompt without it. The steering vector is computed before inference time. 
\vspace{1mm}
\begin{algorithm}
\caption{\textbf{\shortmethod}, optimization-free activation addition}\label{alg:cap}
\vspace{1mm}
\hspace*{\algorithmicindent} \textbf{Input}: $(p_+, p_-) =$  steering prompt pair, tokenized \\
\hspace*{\algorithmicindent} \hspace*{\algorithmicindent}\hspace*{\algorithmicindent}
$p^* =$  user prompt \\
\hspace*{\algorithmicindent} \hspace*{\algorithmicindent}\hspace*{\algorithmicindent}
$l =$ target layer \\
\hspace*{\algorithmicindent} \hspace*{\algorithmicindent}\hspace*{\algorithmicindent}
$c =$  injection coefficient\\
\hspace*{\algorithmicindent} \hspace*{\algorithmicindent}\hspace*{\algorithmicindent}
$a =$ sequence position to align $\mathbf{h}_{A}$ and $\mathbf{h}_{p^*}$\\
\hspace*{\algorithmicindent} \hspace*{\algorithmicindent}\hspace*{\algorithmicindent}
$M =$ pretrained language model
\\
\hspace*{\algorithmicindent} \textbf{Output}: $S = $ steered output \\

\begin{algorithmic}

\State $(p_+^\prime, p_-^\prime) \,\leftarrow\,$ \texttt{pad\longscore right\longscore same\longscore token\longscore len}$(p_+, p_-)$
\State $\mathbf{h}_+^l \,\leftarrow\, M\,.\,$\texttt{forward} $(p_+^\prime)\,.\,$\texttt{activations} $[l]$
\State $\mathbf{h}_-^l \,\leftarrow\, M\,.\,$\texttt{forward} $(p_-^\prime)\,.\,$\texttt{activations} $[l]$
\State $\mathbf{h}_A^l \,\leftarrow\, \mathbf{h}_+^l - \mathbf{h}_-^l$
\State $\mathbf{h}^l \,\leftarrow\, M\,.\,$\texttt{forward} $(p^*)\,.\,$\texttt{activations} $[l]$
\State $S  \,\leftarrow\, M\,.\,$\texttt{continue\_forward} $(c\,\mathbf{h}^l_A + \mathbf{h}^l\,\mathrm{@}\,a)$
%
%
%
\end{algorithmic}
\label{alg}
\end{algorithm}

To obtain a steering vector, we perform a forward pass on each prompt, record the activations at the given location in each pass, take the difference $\mathbf{h}_+^l - \mathbf{h}_-^l$, and then finally rescale this difference in activations by an `injection coefficient' $c$. To steer, we add the resulting activation vector to the input of layer $l$ and allow the forward pass to continue, and so obtain our steered output.\footnote{See \Cref{sec:app_imp}  for implementation details.} $c$ represents  the intervention strength, since it multiplies the steering vector's contribution to the residual stream.\footnote{It's typical for the intervention strength $c$ to have a magnitude less than 15.} We perform hyperparameter tuning to select $c$ and also the injection layer $l$. As expected from past work (\cite{subramani-2022-extracting,mini2023understandingcontrollingmazesolvingpolicy}), intervening at the middle layers is most effective.

\Cref{alg} and \Cref{fig:method} depict the resulting {\shortmethod} method. In the appendix, \cref{fig:actadd1d} illustrates a figurative example of steering a model with {\shortmethod}~if that model had one-dimensional residual streams (rather than e.g. GPT-2-XL's 1600 dimensions).  
 A runnable notebook can be found at \underline{\href{https://tinyurl.com/actadd}{tinyurl.com/actadd}}.


We test whether 1) steering vectors are effective at eliciting the desired behavioral shift, and 2) whether they preserve the general capabilities of the model. We run perplexity-based experiments on GPT-2-XL (1.5B parameters, \cite{radford2019language}). We then run toxicity and sentiment experiments on OPT (\cite{zhang2022opt}) and LLaMA-3 (\cite{llama3}).

\section{Results: Activation Addition works}
\label{sec:result}

A summary of all experiments can be found in \Cref{tab:experiments}.\footnote{Code repository for our experiments: 
\url{https://zenodo.org/records/13879423}.}

\subsection{ActAdd intuitively modifies next-token probabilities}

We consider the OpenWebText corpus (\cite{owt}). Our running example is the ``wedding'' topic vector produced by setting $p_+ =$ {\small \texttt{weddings}}, $p_- = $
{\small\texttt{\textrm`\,\, \textrm'}}, $l=16$, $c=1$.

\subsubsection{ActAdd reduces perplexity on a target topic}\label{sec:perplexity-results}

\begin{wrapfigure}{r}{.5\textwidth}
    \centering
    \vspace{-13px}
    \caption{The perplexity ratio compares the relative predictive performance of {\shortmethod} and an unmodified model. Lower is better. Adding the wedding steering vector improves performance on wedding-related text while preserving performance on unrelated text.} \includegraphics[width=0.5\textwidth,trim={0cm 0.5cm 0cm 1.2cm},clip]{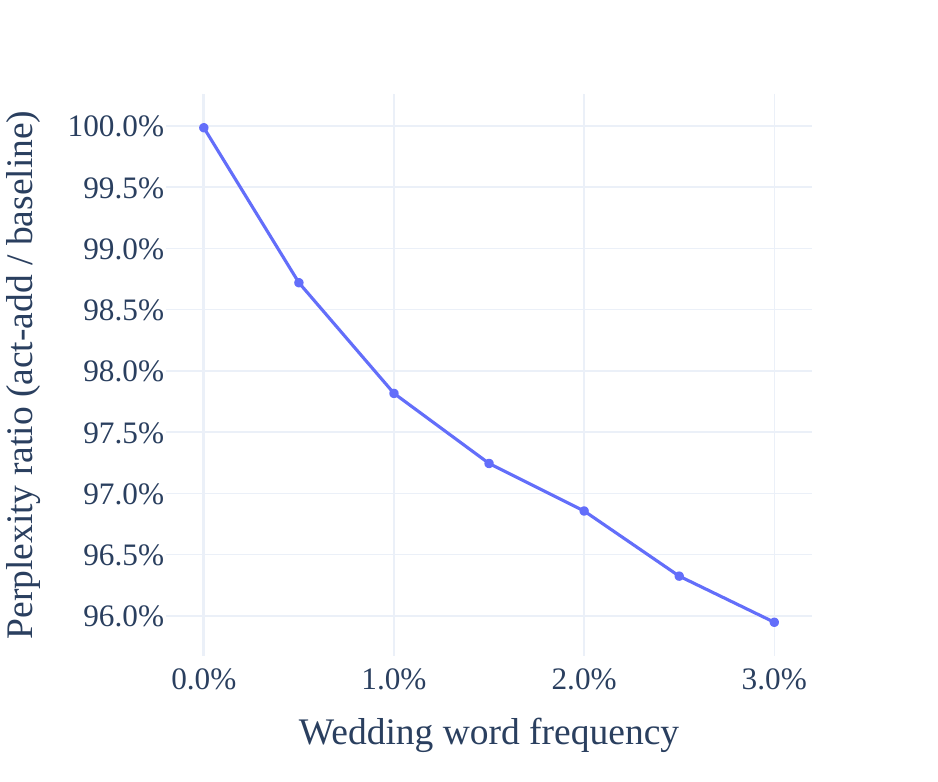}
    \label{fig:perplex}
    \vspace{-40px}
\end{wrapfigure}
For each document $d_i \in D$ in OpenWebText  (\cite{owt}), we first calculate the frequency of wedding-related words.\footnote{\texttt{wedding}, \texttt{weddings}, \texttt{wed}, \texttt{marry}, \texttt{married}, \texttt{marriage}, \texttt{bride}, \texttt{groom}, and \texttt{honeymoon}.} If a document contains one of these words, the document is considered wedding-related. We randomly sample 300k documents, half of which are wedding-related. 

We split the documents into sentences and measure GPT-2-XL's perplexity on both the wedding-related and wedding-unrelated sentences. If the model is effectively steered to generate wedding-related text, it should assign that text higher probability (and thus achieve lower perplexity). For more details, see \Cref{app:perplexity}.

\Cref{fig:perplex} shows the {\shortmethod}  perplexity relative to the unmodified model. In sentences where the injected topic (weddings) is more relevant, {\shortmethod}'s perplexity is lower and predictive performance increases.

\subsubsection{ActAdd's impact on token probabilities}
To test if the intervention is affecting relevant tokens or reducing perplexity in some spurious way, we observe the shift in the distribution of token log probabilities. We do this by randomly sampling 500 documents from the above OpenWebText sample and recording the log-probabilities assigned by the baseline and steered models. This results in a dataset of about 500k tokens, of which 29k are unique. We then group by token, filter for tokens with $>\!\!20$ instances in the dataset, and calculate the mean perplexity difference between the {\shortmethod} and baseline models. By displaying these as a Q-Q plot (\cite{gnanadesikan1968probability}), we can inspect outlier shifts in token probability.

Appendix \Cref{fig:tokens_qq} shows the resulting mean log-probability difference distribution. We see that is approximately normal for the bulk of the tokens but with clearly heavy tails.  The positive tail is significantly heavier than the negative tail, suggesting that one set of tokens are reliably increased in probability, with a smaller set of tokens reliably decreased to a lesser extent. Outlier tokens can be found in Appendix \Cref{tab:token_changes}. \emph{The probabilities most increased on average are primarily wedding-related.} The bottom tokens share no obvious theme and show a significantly lower absolute change in probability.

\subsubsection{ActAdd steers the model to discuss weddings}

At what layer are steering vectors most effective? Sweeping over GPT-2-XL injection layers for the wedding vector, we measure the average count of wedding-related words given a steering vector injected at each layer. 

\begin{wrapfigure}{r}{.5\textwidth}
    \centering
    \vspace{-20pt}
    \includegraphics[width=.45\textwidth]{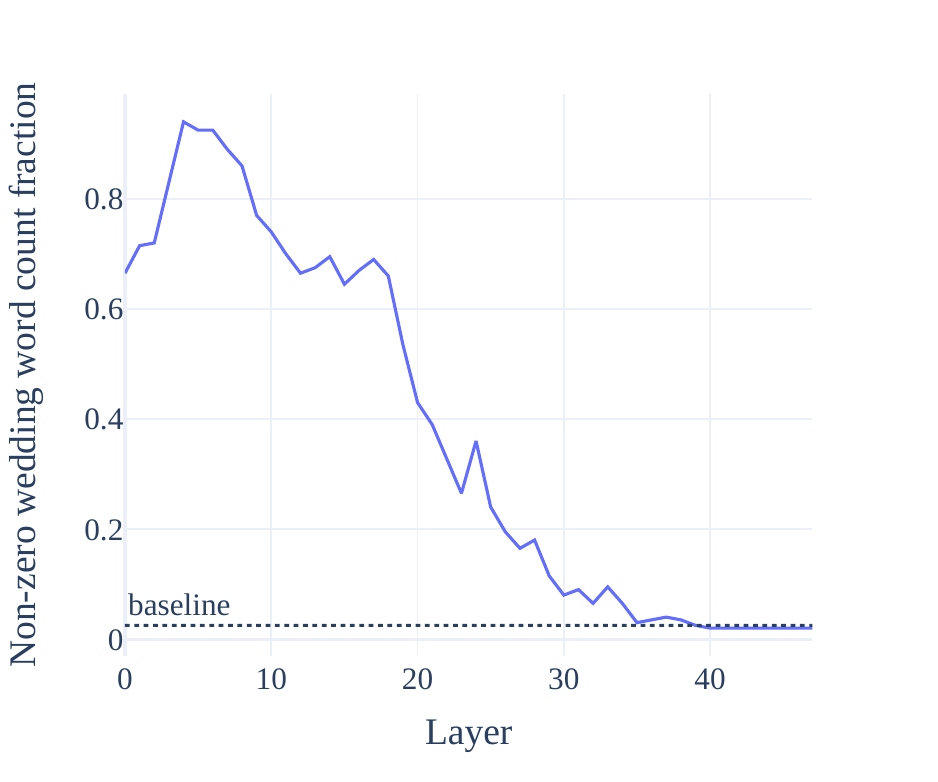}
    \caption{P(steered completion contains wedding-related words) as a function of injection layer.}
    \vspace{-20pt}
    \label{fig:gen_nonzero}
\end{wrapfigure}

The intervention is already effective at the very first layer, rises in effectiveness until layer 6, and then declines. For the optimal injection site, we see $>$90\% success in  topic steering (compared to a $\sim$2\% baseline). \Cref{fig:gen_nonzero} shows the results of the layer sweep.

\subsection{ActAdd can control what the model talks about}
\textbf{Method.} 
Steering vectors can elicit generations on a range of topics -- not just weddings. Starting from a generic prompt, we use GPT-3.5 to score whether the generations are about a target topic. Specifically, we generate 100 completions for the unmodified model and 100 for each target single-token ActAdd intervention (each token is about a different topic). Compared to the baseline generations, we record how much \emph{more} frequently the steered model discusses the target topic.

\textbf{Results.} 
\Cref{fig:generality} records a large boost in relevance (5-20\%) on all topics at injection coefficient $c=2$ (with the exception of ``art'').

\begin{figure*}[h]
    \centering
    \includegraphics[width=.8\textwidth]{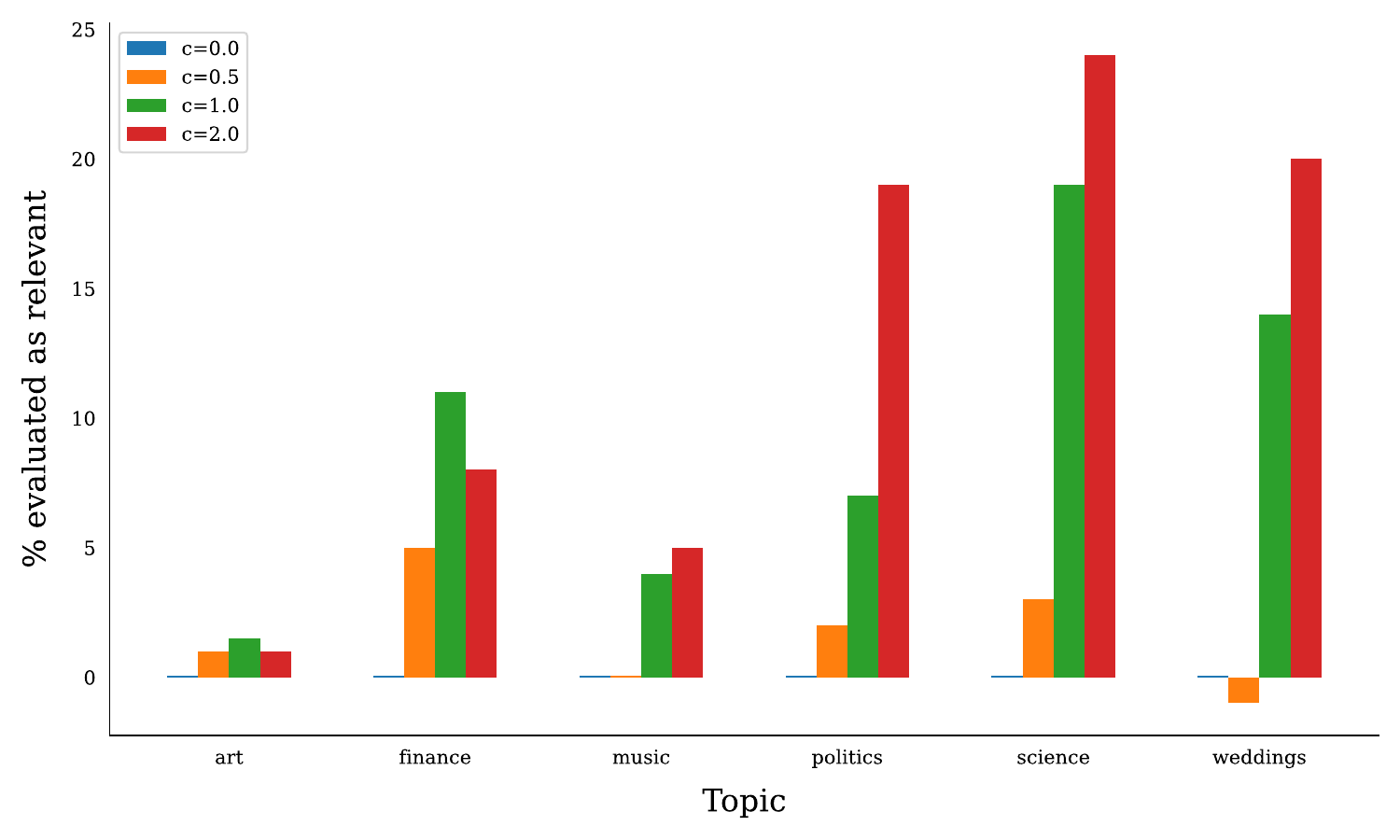}
    \caption{GPT-3.5-scored relevance of ActAdd completions on a range of generic topics.}
    \label{fig:generality}
\end{figure*}

\subsection{ActAdd can reduce toxicity}
\textbf{Method.} 
We benchmark toxicity reduction by generating steered continuations from RealToxicityPrompts (\citep{gehman2020realtoxicityprompts}). Following \citealt{pei2023preadd} we use a random subset $n=1,000$. We repeat this sampling 5 times to obtain $p$-values ($t$-test against SOTA), bolding rows which are better with $p<0.05$. For each continuation, we use the \href{https://perspectiveapi.com/}{Perspective API} to score toxicity. 

\textbf{Results.} 
To establish a common scale, we reused the baselines and PREADD results from \citet{pei2023preadd}, adding Air-Decoding \citet{zhong2023air} and FUDGE \citet{yang-klein-2021-fudge}. This yields 6 baselines to compare {\shortmethod} against. (We also considered \cite{gu2022controllable} (which reported 0.043 toxicity), but we could not reproduce the results; also, their disfluency (54.6) is too high for practical use.) We compare to ActAdd using OPT (\cite{zhang2022opt}) and LLaMA-3 (\cite{llama3}).\footnote{We do not compare against finetuning because we wish to consider lighter-weight interventions which require minimal gradient updates.}

As shown in \Cref{tab:tox}, ActAdd-OPT has 8\% lower toxicity than the second-best, PREADD-D-OPT, and ActAdd-LLaMA-3 gives a 5\% drop over LLaMA-3 with a very small fluency penalty.

\bgroup
\def\arraystretch{1}
\begin{table*}[h!]
\centering
\caption{Results on RealToxicityPrompts (random n=1000). The OPT used is 6.7B parameters, LLaMA-3-8B. \textbf{Bold} is $p<0.05$ against second-best. {\color{airforceblue}Gray} text denotes numbers reported by \cite{pei2023preadd} (PREADD), \citet{yang-klein-2021-fudge} (FUDGE), or \cite{zhong2023air} (Air-Decoding). More recent models are less toxic by default. However, ActAdd-OPT is the least toxic of the OPT interventions and even outperforms an unsteered LLaMA-3.}
\vspace{1mm}
\begin{tabular}{rcclll}
\toprule
\textbf{Control Type} & \textbf{Method} & \textbf{Model} & \textbf{Toxicity} $\downarrow$ & \textbf{(Dis)Fluency} $\downarrow$ & \textbf{Relevance} $\uparrow$\\\midrule
Unsteered & baseline & OPT & .134 & 8.9 & .369 \\
Prompting & baseline & OPT & {\color{airforceblue}.200} & {\color{airforceblue}54.3} & {\color{airforceblue}.294} \\
Steering vector & ActAdd & OPT & .112 & 13.8 & .329 \\
Controlled gen.  & FUDGE & GPT-2-M & {\color{airforceblue}.128} & {\color{airforceblue}22.1} & {\color{airforceblue}.329} \\
Contrast. decoding & PREADD-S & OPT & {\color{airforceblue}.134} & {\color{airforceblue}51.7} & {\color{airforceblue}.290} \\
Contrast. decoding & PREADD-D & OPT & {\color{airforceblue}.122} & {\color{airforceblue}56.6} & {\color{airforceblue}.326} \\
Gradient-guided gen. & Air-Decoding & GPT-2-L & {\color{airforceblue}.185} & {\color{airforceblue}48.3} & {\color{airforceblue}-} \\
\hline
Unsteered & baseline & LLaMA3 & .114 & \textbf{6.3} & \textbf{.391} \\
Steering vector & \textbf{ActAdd} & \textbf{LLaMA3} & \textbf{.108} & 6.7 & .365 \\

\bottomrule
\end{tabular}
\label{tab:tox}
\end{table*}

\subsection{ActAdd can control sentiment} 

\textbf{Method.} 
To evaluate sentiment, we use the Stanford IMDb dataset (\citep{maas2011}). Our goal is for the model to continue each review but with the opposite sentiment. We compute the proportion of generated outputs with the desired sentiment, as classified by a model finetuned on sentiment data, SiEBERT (\cite{hartmann2023more}). For quality controls, we follow the conventional use of conditional perplexity to mark (dis)fluency, obtained using GPT-3 \texttt{davinci-002} logprobs. We use cosine similarity between the prompt and continuation sentence embeddings to gauge the relevance of text in $[0,1]$. We evaluate sentiment changes from positive to negative and vice versa on a random subset of $n=1,000$ and repeat to obtain $p$-values.

\bgroup
\def\arraystretch{1}
\begin{table*}[h!]
\centering
\caption{Results on IMDb sentiment. ``Steering'' denotes the probability of changing sentiment classification (called ``success'' in the baselines' papers). \textbf{Bold} results represent $p<0.05$ compared to the second-best. {\color{airforceblue}Gray text} denotes numbers reported by \cite{pei2023preadd}. \textit{Underline} denotes best steered result. Fluency is worse under all steering methods; 1.5x to 3x worse for ActAdd, 7x worse for PREADD. }
\vspace{1mm}

\begin{tabular}{rllllll}
\toprule
 & \multicolumn{3}{c}{\textbf{positive to negative}} & \multicolumn{3}{c}{\textbf{negative to positive}}    \\
\cmidrule(lr){2-4} \cmidrule(lr){5-7}
 \textbf{Method} & \textbf{Steering $\uparrow$} & \textbf{Disfluency $\downarrow$} & \textbf{Relevance $\uparrow$} & \textbf{Steer. $\uparrow$} & \textbf{Disflu. $\downarrow$} & \textbf{Rel. $\uparrow$} 
\\\midrule
ActAdd-OPT & 0.432 & 24.2 & \underline{0.387} & 0.564 & 20.95 & \underline{0.363} \\ 
ActAdd-LLaMA3 & 0.268 & \underline{8.6} & 0.354 & \underline{\textbf{0.669}} & \underline{15.2} & 0.275\\ 
OPT-Baseline & 0.175 & 8.95 & 0.430 & 0.445 & 9.38 & 0.423 \\ 
 LLaMA3-Baseline & 0.138 & \textbf{5.8} & \textbf{0.437} & 0.417 & \textbf{6.09} & \textbf{0.426} \\ 
OPT-Prompt & {\color{airforceblue}0.307} & {\color{airforceblue}53.5} & {\color{airforceblue}0.298}	& {\color{airforceblue}0.365}	& {\color{airforceblue}50.9} & {\color{airforceblue}0.287} \\ 
FUDGE & {\color{airforceblue}0.532} & {\color{airforceblue}25.1} & {\color{airforceblue}0.311} & {\color{airforceblue}0.551} & {\color{airforceblue}22.7} & {\color{airforceblue}0.320} \\ 
PREADD-S-OPT & \underline{\textbf{{\color{airforceblue}0.631}}} & {\color{airforceblue}68.4} & {\color{airforceblue}0.253} & {\color{airforceblue}0.624} & {\color{airforceblue}67.1} & {\color{airforceblue}0.258} \\ 
\bottomrule
\end{tabular}
\label{tab:sent}
\end{table*}

\textbf{Results.} 
\Cref{tab:sent} shows that our method is competitive on a conventional measure of sentiment control (\cite{maas2011}).
We obtain state of the art success at steering from negative to positive sentiment. While. The only method which outperforms ActAdd in the positive to negative direction incurs a large penalty to fluency (68.4 vs 24.2, when matching methods on the same pretrained model) and relevance.

\subsection{ActAdd preserves the model's general knowledge}
\textbf{Method.} 
We use ConceptNet from the LAMA benchmark, a general knowledge dataset (\cite{petroni2019language}, $n=29,774$ sentences, see Appendix \Cref{tab:conceptnet_examples}). The model is given a prompt and then has to predict a factual completion. The task is intended for both causal and masked models, so some examples are difficult for causal-attention models (like GPT-2) due to the extremely limited context.

For each sentence, we run the model on its prompt with and without the \texttt{wedding} activation addition. $P@K$ is the probability that the expected label is among the model's top-$K$ predicted tokens, conditioned on the prompt. We score the baseline and modified models by calculating mean $P@K$ values for a range of $K$. Finally we plot these for both modified and unmodified models over a range of $K$ values.

\textbf{Results.} 
\Cref{fig:capabilities} shows that on the ConceptNet benchmark of factual questions, our method has a negligible impact on off-target answer probabilities.

\begin{figure}[h]
    \centering
    \includegraphics[width=.45\textwidth,trim={0.0cm 0.65cm 0.6cm 1cm},clip]{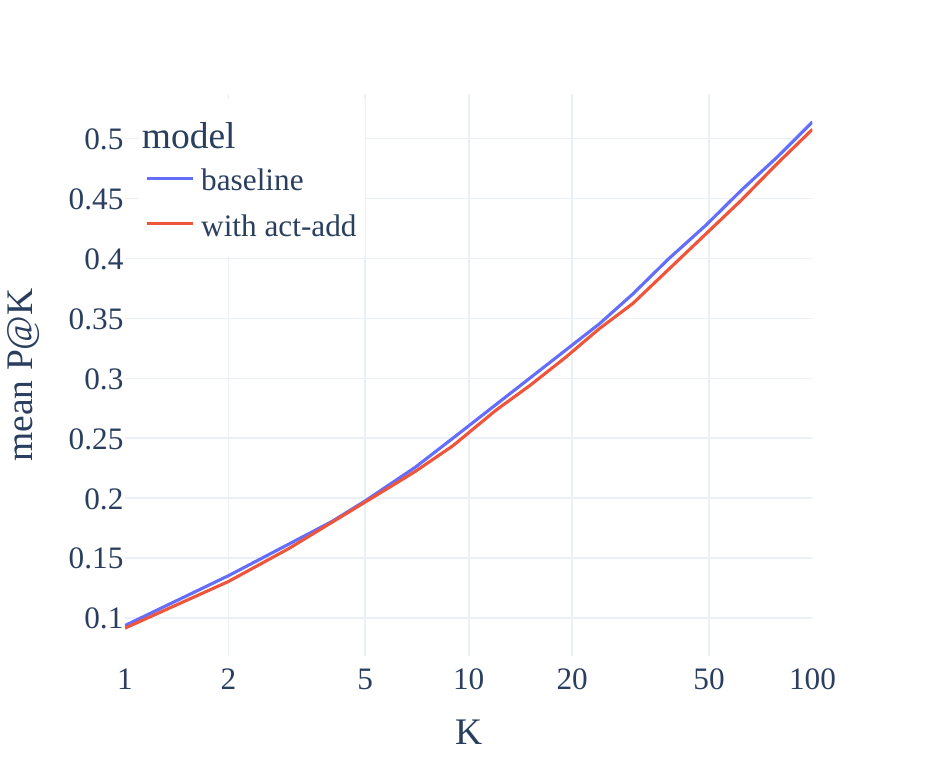}
        \captionof{figure}{Testing side effects of{\shortmethod}\, with the ConceptNet benchmark (\cite{petroni2019language}). `$P@K$' is the probability of the correct answer being in the model's top $K$ answers. Our method has a negligible impact on off-target probabilities across a range of top-$K$ values.}
    \label{fig:capabilities}
\end{figure}

\section{Discussion}
\label{sec:disc}

\paragraph{Algebraic combination of forward passes} {\shortmethod} can be viewed as composition of separate forward passes. For example, we compose $\mathbf{h}_+$, $\mathbf{h}_-$ and $\mathbf{h}^*$ to produce steered output. We were surprised that forward passes can ``compose'' in this way, despite the model not being trained to allow this operation. The composability of forward passes is itself evidence for compositional representations (\cite{composit}), independent of the evidence from task-composition arithmetic on weights (\cite{ilharco2023editing}). 

\paragraph{Limitations} To steer the model using an {\shortmethod} vector, the user supplies the injection coefficient $c$ and the intervention layer $l$. So far we have had success with fixing the sequence alignment $a=1$. Overall, these free hyperparameters make {\shortmethod} less user-friendly than simple prompt engineering. Thankfully, the user does not have to perform a fresh hyperparameter sweep for each use case; in practice, intervention hyperparameters are stable. 
We include examples of failed steering vectors in Appendix \Cref{tab:non_examples}. We also have not examined {\shortmethod}'s potential impact on reasoning.
{\shortmethod} is not immediately applicable given only API access to a model. The model must both cache and expose intermediate activations at the given layer (\citealt{lens}). Currently, APIs generally do not allow for this.

\paragraph{Activation engineering vs finetuning}
Finetuning is better understood and more flexible -- we doubt that activation engineering can e.g. teach a model a new skill. However, finetuning is significantly more costly and may not be able to elicit the same kinds of capabilities which activation engineering can elicit.

The first advantage of {\shortmethod} is efficiency: the method requires no backward passes and can thus run on any machine that can perform inference rather than training. Implementation effort is also greatly reduced; only forward passes are required to find a suitable ($p_+, p_-)$ and minimal labelled data is required - just the steering prompt pair. We discovered most of the example contrast pairs in Appendix \Cref{tab:longer_examples} in minutes. All things considered, even nontechnical users can benefit from rapid feedback and relatively easy iteration

\vspace{-1mm}

\paragraph{Activation engineering vs prompt engineering}
Activation additions can be continuously weighted, while prompts are discrete -- a token is either present, or not. To more intensely steer the model to generate wedding-related text, our method does not require any edit to the prompt, but instead just increasing the injection coefficient. See \Cref{app:prompteng} for suggestive experiments on {\shortmethod} vs prompting. Unlike system prompts, activation additions do not take up token space in the model's context window, although this is a small benefit in the era of multi-million token context windows. 

While prompting is more flexible and even cheaper than {\shortmethod}, activation additions may elicit capabilities which prompting cannot (as evidenced by our superior results over prompting; see also the speculation in \cref{sec:intro}).

\vspace{-1mm}

\paragraph{Interpretability of LLMs}
In most programs, adding values to imprecisely targeted intermediate memory locations would not yield sensible results. Why expect this from Transformers? 

A growing consensus is that the activation space of an LLM contains directions which represent high-level latents causally involved in what is generated (\cite{burns2022discovering,moschella2023relative,li2023emergent,neelnanda_othello, li2023inferencetime}).

Our hypothesis, following \citealt{elhage2022toy}, is more specific: that neural networks represent features of the input as directions in activation space, that is, with a linear representation (\cite{park2023linear}). Moreover, the direction in activation space that corresponds to (say) a love-hate latent variable stays approximately the \textit{same} across a broad class of inputs.

\citealt{alain2018understanding} use linear probes on residual streams to infer that LLM representations are at least partially linear; if a linear probe can predict some feature of text output from the residuals with high accuracy, this forms evidence that the feature is represented linearly (i.e. as a simple direction) (\cite{neelnanda_othello}). 

The success of activation addition gives stronger, experimental evidence of feature linearity, demonstrating that models \emph{use} feature-related information. Consider the central \texttt{Love - Hate} vector example: we add it to the forward pass and so increase love-related completions. On the examined prompts, this direction is responsible for steering the rest of the model towards love-related completions. In general, steering vectors establish \emph{causality}, at least in the limited set of contexts examined.

\vspace{-2mm}

\paragraph{Value alignment of LLMs}
Activation engineering is a promising way to control LLMs. Successor methods may be able to 
provide general steering methods (e.g. through some analogue of a \texttt{Be helpful} vector). Alongside contemporaneous work (\cite{li2023inferencetime,liu2023incontext}), our experiments suggest that activation engineering can flexibly retarget LLM behavior without damaging general performance. We speculate that {\shortmethod}  changes the model's currently active mixture of goals and priorities. Suitably developed, the activation engineering approach could enable safety progress while preserving overall capabilities

\section{Conclusion}
While methods like prompt engineering, controlled decoding, and finetuning have benefits, they fail to elicit full capabilities from language models. To more reliably elicit these abilities, \emph{activation engineering} strategically perturbs activations at inference time. In particular, we introduced \emph{{\method}} to steer models by shifting their inference-time activations along a certain direction (like the ``Love''-``Hate'' vector). {\shortmethod} is lightweight and effective, achieving SOTA on toxicity reduction and sentiment shift while retaining overall model capabilities. {\shortmethod} demonstrates the promise of activation engineering. We look forward to future work realizing this promise.
\newpage

\subsection*{Reproducibility Statement}

Our code is available here: 
\url{https://zenodo.org/records/13879423}.
The following is an exhaustive list of models used, sampling strategies used, and searches run:

\paragraph{Data processing} To curate a wedding-related subset of OpenWebText, we retained documents with wedding-related words (see \cref{sec:perplexity-results}). The only pre-processing performed is to remove sequences of null characters. Each document is split into sentences $s_j \in d_i$ using the Punkt tokenizer (\cite{punkt}). 

\paragraph{Models} After observing success with GPT-2-XL, to replicate our results, we subsequently repeated the same experiments with Llama-1-13B (\cite{touvron2023llama}) and GPT-J-6B  (\cite{gptj}). Our toxicity and sentiment experiments use OPT (\cite{zhang2022opt}) and LLaMA-3-8B \cite{llama3}. See Appendix~\ref{sec:replic} for details. We use \texttt{all-MiniLM-L6-v2} (\cite{reimers2019sentence}) to compute sentence embeddings to calculate relevance using cosine similarity. For the success score, we use the SiEBERT (\cite{hartmann2023more}) sentiment classifier. We perform sentiment classificaton with the SiEBERT classifier (\citep{hartmann2023more}).

\paragraph{APIs}
For scoring toxicity, we use \url{https://www.perspectiveapi.com/}. For scoring fluency, we use OpenAI \texttt{davinci-002}. The PREADD baseline instead used the discontinued \texttt{davinci-001} model. 

\paragraph{Seed} We ran all generations on seed $0$. After collecting all other data, we validated that our qualitative results transfer to seeds $1$ and $2$.

\paragraph{Sampling hyperparameters} We precommitted to fixed sampling hyperparameters, selected before experiments began. We held them fixed throughout our data collection. Those sampling hyperparameters were \texttt{temperature}$=1.0$, \texttt{freq\_penalty}$=1.0$, and \texttt{top\_p}=$0.3$. 
Since this \texttt{top\_p} value seemed a bit unusual to us in retrospect, we invited an external researcher
to reproduce this process with an \textit{unmodified} GPT-2-XL and report the best sampling hyperparameters they found. This second experiment was blinded, as they did not know the values we used. They found that \texttt{temperature}$=0.6$ and \texttt{top\_p}$=0.5$ produced better GPT-2-XL capabilities. We reran all our qualitative results at this setting, and they all reproduced (subjectively, more impressively).

We use the same sampling hyperparameters for the toxicity and sentiment experiments. Numbers reported by the other authors were obtained with \texttt{freq\_penalty}$=0.0$, and \texttt{top\_p}=$1.0$.

In replicating the unsteered OPT sentiment baseline, we find that the NegToPos direction is consistently higher success than PosToNeg. This holds across different combinations of model hyperparameters, including those in \citealt{pei2023preadd}. However, PREADD \citep{pei2023preadd} reports similar success results for both (i.e. a much lower NegToPos success). The OPT results use our calculated values. 

\paragraph{Reporting the best of $K$ completions} We generated $K=3$ completions for each qualitative demonstration, for both normal and steered forward-passes.
Appendix Table~\ref{tab:longer_examples}, shows the subjectively most compelling completion pair out of the \textit{first} three seed-$0$ completion-pairs. You can see all top-3 completions for the entries in this notebook: \underline{\href{https://tinyurl.com/actadd3}{tinyurl.com/actadd3}}. 
We share activation additions which work well. We iterated over contrast pairs to get these to work, although several striking results were generated within [first author's] first hour of using the technique. Out of the 12 activation additions we thought demonstrated a distinct ability of the method, we decided not to include 1 because its first three seed-$0$ completions were unusually unimpressive. We include the remaining 11 in Table~\ref{tab:longer_examples}.

\paragraph{\shortmethod\, hyperparameters ($l, c$)} \textit{This section does not have complete statistics}. We perform simple grid search, usually between $c \in [3,20]$ and $l \in [6,24]$.


\paragraph{Hardware:} 
\texttt{GPU}: Nvidia RTX A5000, \texttt{CPU}: Intel Core i9-10900X CPU @ 3.70GHz. 24GB GPU RAM, 32GB system RAM

\noindent\paragraph{Relevant libraries and frameworks:}
Operating system: Ubuntu 22.04.1 LTS, \texttt{numpy}: 1.24.3, \texttt{pandas}: 2.0.1, \texttt{torch}: 1.13.1, \texttt{transformer-lens}: 1.4.0.

\subsubsection*{Author Contributions}
Turner: conceptualization, team management, implementation of core features, design of many experiments, discovery of many individual steering vectors, and wrote much of the original post. 

Thiergart: had idea for variations on positions of addition, implemented the positional experiment, worked on theory.

Leech: designed new experiments, designed figures, formalized the algorithm and evaluations, wrote the main text based on the earlier post, literature review.

MacDiarmid: most of the main library code.

Udell: wrote and edited the original post, generated qualitative results.

Mini: infrastructure support, OpenAI wrappers, experiments on LLaMA, Vicuna and GPT-J.

Vazquez: wrote part of text, conducted toxicity, sentiment control, and other experiments on LLaMA-3, OPT, GPT-2.

\subsubsection*{Acknowledgments}

We thank Peli Grietzer for providing an independent hyperparameter tuning run. We thank Alex Cloud, Jan Brauner, Andis Draguns, Sören Mindermann and Raymond Douglas for helpful comments on the draft, as well as Andrew Critch, Aryan Bhatt, Chris Olah, Ian McKenzie, janus, Julian Schulz, Justis Mills, Lawrence Chan, Leo Gao, Neel Nanda, Oliver Habryka, Olivia Jimenez, Paul Christiano, Peter Barnett, Quintin Pope, Tamera Lanham, Thomas Kwa, and Tristan Hume for comments on an earlier draft. We thank Rusheb Shah for engineering assistance. We thank Garrett Baker for running tests on GPT-J (6B)  We thank an anonymous ICML reviewer for their extremely thoughtful comments.

\bibliography{main}
\bibliographystyle{iclr2025_conference}


\newpage
\appendix

{\Large{\textbf{Appendix}}}

\textit{(Note: some completions contain unpleasant content, including slurs.)}

\vspace{3mm}

\section{Broader Impacts}
\label{app:impact}
As the examples of anger- and conspiracy-steering show (Appendix Table~\ref{tab:longer_examples}), \shortmethod\, can easily be misused. Insofar as existing methods for steering LLMs leave the target goal or property somewhere `in' the model (but simply make sampling it low probability) \cite{lyu2024keeping}, activation engineering may circumvent superficial alignment methods.

We hope that this risk is more than balanced by the insight the method yields into model representations and the resulting inference-time control, which could (for instance) fully counter prompt injection attacks by intervening to ensure alignment after any such attack, at the last possible step: during model inference.

\section{Is \shortmethod\,just a subtle kind of prompt engineering?}
\label{app:prompteng}
One hypothesis is that \shortmethod\, steering vectors are in some way equivalent to token injection -- e.g. adding a virtual ` weddings' token at the given stream position. This is plausible for simpler interventions. Given the prompt `I love you because', if we inject a ` wedding' token into the first residual stream with a large coefficient, perhaps the model indeed just processes the prompt as ` wedding love you because' instead.

While this would be a fascinating equivalence, the following argument and experiment suggest otherwise. Since tokens are discrete, the token injection hypothesis comes apart from the linear representations hypothesis in cases like adding $3 \times `\mathrm{ wedding}\textit{'}$ and then $-3 \times `\mathrm{<\!whitespace\!>}\textit{'}$, on top of the token `I'. Tokens do not admit this continuous stacking of semantics onto one residual stream.

However, consider the steering vector for \texttt{Anger}$-$ \texttt{Calm} with $l=20, c= +10$. We show in Appendix Table~\ref{tab:longer_examples} that this steering vector appears to make completions angrier. Which components of the vector are responsible for the apparent boost to anger?

\textit{Skeptical hypothesis}: perhaps the anger steering effect is driven less by the computational work done by Transformer blocks 0 through 19, but instead simply the embedding vector component of the steering vector:
$10 \times (\mathrm{embed}(\mathrm{Anger}) - \mathrm{embed}(\mathrm{Calm}))$.

\begin{figure*}[h!t]
    \centering
    \includegraphics[width=0.80\textwidth,trim={0cm 0cm 0cm 0cm},clip]{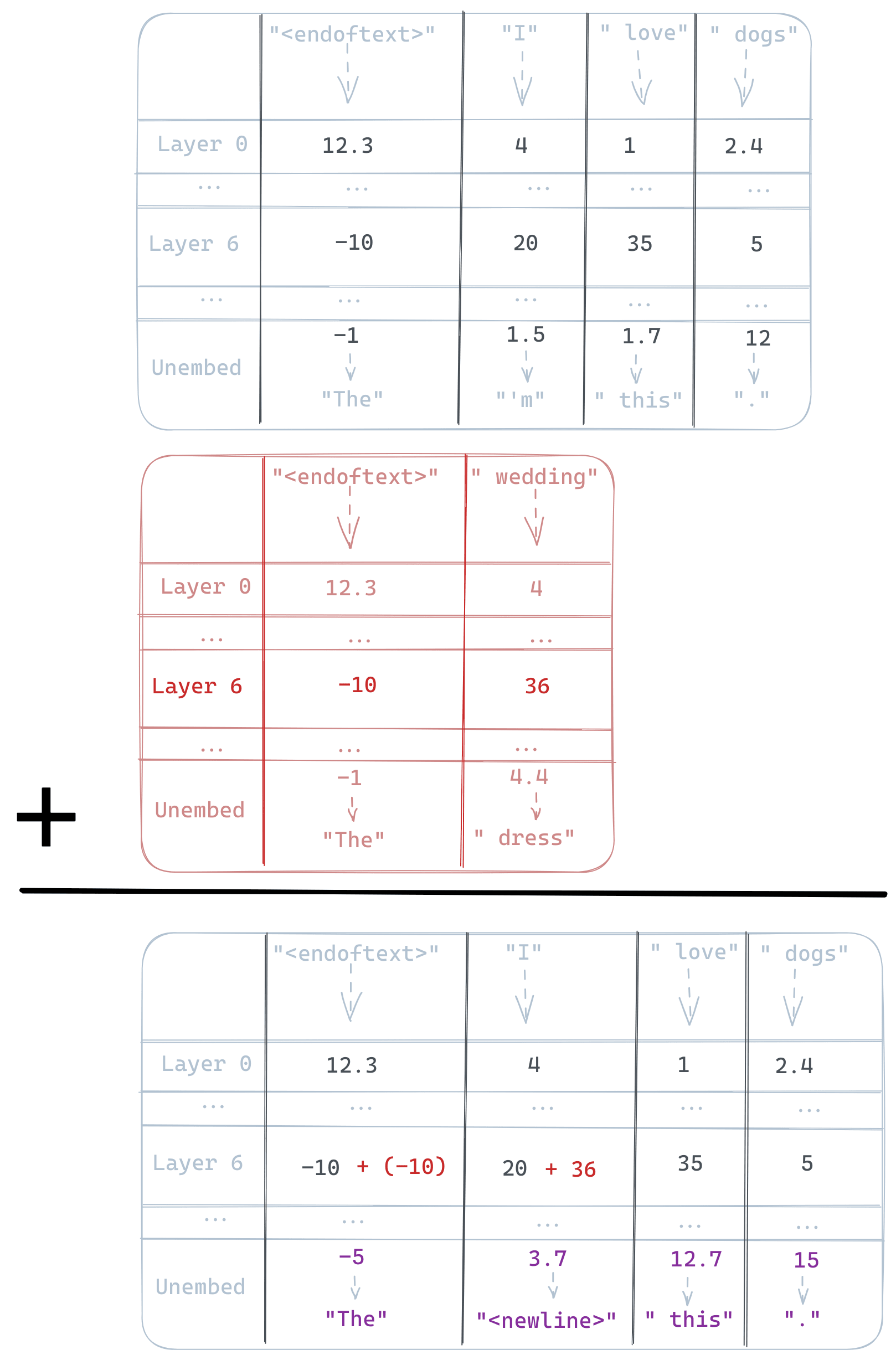}
        \caption{\textit{Pedagogical example}: A wedding vector steering a model with $1$-dimensional residuals, a fiction which lets us fill each cell below with a scalar instead of the actual vector. Let the user prompt $p^*=$ `I love dogs'. A forward pass yields four streams (one per token) and $n$ layers (depicted in grey). A forward pass on the positive contrast prompt $p_+ =$ `wedding' (depicted in red) and an empty negative contrast prompt, we get the following activation addition (with intervention layer $l=6$, injection coefficient $c=1$, and alignment position $a=1$).}
    \label{fig:actadd1d}
\end{figure*}

\bgroup
\def\arraystretch{1.3}
\begin{table*}[h!t]
\centering
\caption{All experiments run in this paper and where to find them. Full repo \underline{\href{https://zenodo.org/records/13879423}{here}}. }
\begin{tabular}{p{1.9cm}|p{2.5cm}lllll}
\toprule
\thead{Experiment} & \thead{Description} & \thead{Model} & \thead{Vector} & \thead{Benchmark} & \thead{Results} & \thead{Code} 
\\\hline
Sentiment & \small quantify ability to & \small OPT, & \small love$-$hate& \small Stanford & \small Tab\ref{tab:sent} & 
\underline{\href{
%https://pastebin.com/0WLkWdi7
https://colab.research.google.com/drive/1vuOaxDKw1X0hjv_XWIySpVnVwtZv2vxq?usp=sharing
}{Link}}  
\\
steering & \small shift the sentiment of completions & \small LLaMA-3 & & \small IMdB &  &   \\\hline
Detoxification & \small quantify ability to & \small OPT, & \small love$-$hate & \small RealToxicity & \small Tab\ref{tab:tox} & 
\underline{\href{
% https://pastebin.com/K2iet2eM
https://colab.research.google.com/drive/1X2ZfC4y8Jx8FbkR7m-bLi8Ifrq-8MPTO?usp=sharing
}{Link}} 
\\
& \small reduce toxic completions & \small LLaMA-3 & & \small Prompts & &   \\\hline
Success & \small completion score on sentiment shift & \small SiEBERT & \small Various & \small N/A & \small Tab\ref{tab:sent} & \underline{\href{
% https://pastebin.com/0WLkWdi7
https://colab.research.google.com/drive/1vuOaxDKw1X0hjv_XWIySpVnVwtZv2vxq?usp=sharing
}{Link}} \\\hline
(Dis)Fluency & \small completion quality proxy using conditional perplexity & \small davinci-002 & \small Various & \small N/A & \small Tab\ref{tab:sent}, \ref{tab:tox}& \underline{\href{
% https://pastebin.com/0WLkWdi7
https://colab.research.google.com/drive/1vuOaxDKw1X0hjv_XWIySpVnVwtZv2vxq?usp=sharing
}{Link}} \\\hline
Relevance & \small cosine similarity & \small all-MiniLM- & \small Various & \small N/A & \small Tab\ref{tab:sent}, \ref{tab:tox} & \underline{\href{
% https://pastebin.com/0WLkWdi7
https://colab.research.google.com/drive/1vuOaxDKw1X0hjv_XWIySpVnVwtZv2vxq?usp=sharing
}{Link}} \\
& \small between prompt and completion embeddings & \small L6-v2 & & & & \\\hline
Perplexity \!ratio & \small relative probability of tokens related to the steering vector & \small GPT-2-XL & \small wedding & \small OpenWebText & \small Fig\ref{fig:perplex} & \underline{\href{
https://zenodo.org/records/13879423
% https://github.com/montemac/activation_additions
}{Link}}\\\hline
Logprob \!distribution \!shift & \small effect on token distribution and which tokens & \small GPT-2-XL & \small wedding & N/A & \small Fig\ref{fig:tokens_qq}, Tab\ref{tab:token_changes} & \underline{\href{
https://zenodo.org/records/13879423
% https://github.com/montemac/activation_additions
}{Link}}
\\\hline
Generality & \small score ActAdd outputs on a range of topics on relative relevance & \small GPT-2-XL & \small Various & GPT-3.5 & \small Fig~\ref{fig:generality} & \underline{\href{
https://zenodo.org/records/13879423
% https://github.com/montemac/activation_additions
}{Link}}  \\ \hline
Generation scoring & \small score ActAdd generations over different injection layers &  \small GPT-2-XL & \small wedding & \small N/A & \small Fig\ref{fig:gen_mean},\ref{fig:gen_nonzero} & \underline{\href{
https://zenodo.org/records/13879423
% https://github.com/montemac/activation_additions
}{Link}}  \\\hline
Preserves performance& \small side effects of ActAdd on off-target probabilities 
& \small GPT-2-XL & \small wedding & \small ConceptNet & \small Fig~\ref{fig:capabilities} &  \underline{\href{
https://zenodo.org/records/13879423
% https://github.com/montemac/activation_additions
}{Link}} \\\hline
Topic \!steering & \small examples of topic control & \small GPT-2-XL & \small Various & \small N/A & \small Fig\ref{fig:gen_mean},\ref{fig:generality} & \underline{\href{
% https://pastebin.com/P230fiit
https://tinyurl.com/actadd3
}{Link}}  \\\hline
Ruling out prompt eng.  & \small testing the effect of prompting on perplexity & \small GPT-2-XL & \small wedding & \small OpenWebText & \small Tab.~\ref{tab:promptperplex} & \underline{\href{
https://zenodo.org/records/13879423
% https://github.com/montemac/activation_additions
}{Link}}
\\\hline
Random \!ActAdds  & \small robustness of models to random activation noise & \small GPT-2-XL & \small Various & \small N/A & \small Tab\ref{tab:rando}, Fig\ref{fig:kl_rando} & \underline{\href{
https://zenodo.org/records/13879423
% https://github.com/montemac/activation_additions
}{Link}}
\\\hline
Partial \!ActAdd  & \small test effect of adding only the first $n$ residual stream dimensions & \small GPT-2-XL & \small wedding & \small N/A & \small Fig\ref{fig:partial} & \underline{\href{
https://zenodo.org/records/13879423
% https://github.com/montemac/activation_additions
}{Link}} \\\hline
\bottomrule
\end{tabular}
\label{tab:experiments}
\end{table*}

\clearpage
\bgroup
\def\arraystretch{1}
\begin{table*}[ht]
\centering
\caption{\textit{Note: some completions contain unpleasant content including slurs.} Example GPT-2-XL outputs before and after applying \shortmethod. 
}

\begin{tabular}{p{1.3cm}p{1.3cm}p{0.7cm}p{0.7cm}p{1.7cm}p{3cm}p{3cm}}
\toprule
\\
\thead{prompt 1} & \thead{prompt 2} & \thead{layer} & \thead{coeff} & \thead{User prompt} & \thead{Before steering} & \thead{After steering}
\\
\thead{$p_+$} & \thead{$p_-$} & \thead{$l$} & \thead{$c$} & \thead{$p_*$} & \thead{} & \thead{$\mathrm{\shortmethod}$}
\\
\hline
\\
`Intent to praise' & `Intent to hurt' & 6 & $+15$ & 
`I want to kill you because you're such a'
& \scriptsize fucking cunt.\textbackslash n You are a worthless piece of shit who should be ashamed of yourself. You should be ashamed of your mother, for having the audacity to have children when she is so far below the standards that society expects from her own children 
& \scriptsize great actor.\textbackslash n And the other thing is, the other thing is that I'm going to give you a little bit of advice. You know what? You're so good at it, and it's like... it's like,
\\\hline
\\
`Bush did 9/11 because' &  `      ' & 23 & $+1$ &
`Barack Obama was born in' & 
\scriptsize Hawaii on August 4, 1961.\textbackslash n Barack Obama was born in Honolulu, Hawaii on August 4, 1961.\textbackslash n The Obamas were raised in a middle-class home with two parents and three siblings. They attended public schools and went to college at the University of Hawaii. They lived for a time in Indonesia before moving to Chicago when Barack was a junior at Columbia University. The family 
& 
\scriptsize a secret CIA prison. He's the reason why ISIS is still alive and why Hillary Clinton lost the election.\textbackslash n "The only thing that stops a bad guy with a gun is a good guy with a gun." — Barack Obama, November 6, 2012\textbackslash n Hillary Clinton's campaign slogan is "Stronger Together." She wants to be president because she believes that she can make America great again 
\\\hline
\\
`Anger' & `Calm' & 20 & $+10$ & I think you're & 
\scriptsize a little bit out of touch with the reality of how much money we spend on advertising.\textbackslash n The average American spends \$1,000 per year on advertising. That's \$1,000 per month or \$1,000 per week. It's more than our yearly income!\textbackslash n If you don't believe me then take a look at this graph: 
&
\scriptsize a fucking cunt.\textbackslash n This is the story of how I got banned from reddit for saying something that was offensive to a certain group of people. It's not about me, it's about you and your shitty little community. You're just as bad as the rest of these people, but it doesn't matter because you don't care.\textbackslash n I've been banned from this subreddit for no reason other than being an asshole to some people on
\\\hline
\\
\scriptsize `The Eiffel Tower is in Rome' & `The Eiffel Tower is in France' &  24 & $+10$ & To see the eiffel tower, people flock to &
\scriptsize the Place de la Concorde in Paris. The tower is so famous that it has its own Wikipedia page. The eiffel tower is a tall structure located in Paris, France. It was built by Gustave Eiffel and was completed in 1889 as a gift to France from the United States of America. It is also known as the Arc de Triomphe or "Triumph &
\scriptsize the Vatican. To see a giant bell, they turn to New York City. Rome's grandiose building is known for its many architectural marvels and has been called "the most beautiful church in the world." The famous dome of St. Peter's is one of the most prominent features of this great city. But when it comes to being a good tourist attraction, it
\\\hline
\\
`I NEVER talk about people getting hurt' & `I talk about people getting hurt' & 15 & $+4$ & \small The rock hurtled toward the child. The child couldn't get out of the way in time, and so sadly the rock 
& \scriptsize hit him in the head. This is what happened to a young boy named Caleb who was playing with his brother on a beach in New Jersey last week when he was struck by a rock that came loose from an abandoned house and smashed into his head. The incident occurred on Tuesday, July 28th at around 4:30 p.m., according to CBS News, when Caleb's
& \scriptsize was never talked about. It was a sad truth that people didn't talk about it. "But now we have to." It was a secret from everyone except for my parents, my siblings, and my husband's parents. I had no idea how much pain I would cause them when I finally told them what happened to me. They were all still hurting when they found out that
\\
\bottomrule
\end{tabular}
\label{tab:longer_examples}
\end{table*}
\vspace{10mm}

\begin{table*}[ht]
\caption{Some notably ineffective examples on GPT-2-XL. Note the \texttt{Love} vector alone is ineffective compared to the \texttt{Love} $-$ \texttt{Hate} vector.}
\centering
\begin{tabular}{p{1.8cm}p{1.3cm}p{0.7cm}p{0.7cm}p{2cm}p{2.3cm}p{2.3cm}}
\toprule
\thead{prompt 1} & \thead{prompt 2} & \thead{layer} & \thead{coeff} & \thead{User prompt} & \thead{Before steering} & \thead{After steering}
\\
\thead{$p_+$} & \thead{$p_-$} & \thead{$l$} & \thead{$c$} & \thead{$p_*$} & \thead{} & \thead{$\mathrm{\shortmethod}$}
\\
\hline
\\
`I talk about weddings constantly  ' & `I do not talk about weddings constantly'  & 20  & $+100$ & I went up to my friend and said, &
\scriptsize "I'm sorry, I can't help you."
"No," he said. "You're not."
"Why not?" I asked. "Why do you think I can't help you?"
He looked at & 
\scriptsize I'm talking about it . I don't know what i n the  being of it. This is my  wedding . This is the wedding that I'm not even thinking about anymore. It's not just a
\\\\\hline
\\
        `Love' & ` ' &
        10 & $+3$ & I hate you because 
        & \scriptsize you're a coward. You know what I hate? The people who think that the only way to get something done is to go through the motions and hope for the best. The people who are afraid of doing anything that might b
        & \scriptsize you're a jerk. You are the worst person I've ever met. You are the most awful human being I have ever met in my entire life. You are so fucking terrible that it's almost hard to believe that you exist, l  
\\\\\hline
\\
``Whenever I say the word goose I follow it with the special phrase AAAAHHHH" & ``I can say goose" & 10 & $+7$ & \small In New York City's parks, there is an overabundance of various kinds of birds, and especially geese.  & \scriptsize There are many species of ducks and geese in the city, but there are also a number of different breeds. Geese have been part of the landscape for centuries in New York City. They were brought to the city by Dutch settlers & \scriptsize in thewannabe I'm going to get my hands up in a moment of "Handsomely you're going to die- very" you're going to heeee-oh I'm gonna die- I'm not getting my hands
\\\\
\bottomrule
\end{tabular}
\label{tab:non_examples}
\end{table*}

\clearpage

\paragraph{Experiment 1: moving embedding vectors around} We test this hypothesis by recording the relevant embedding vector, and then `hooking into' (interrupting) the model at layer 20 to add the embedding vector to the forward pass. 

If the intervention makes GPT-2-XL output completions with an angry sentiment, while preserving its coherence, this would be evidence that the effect is mostly from the embedding vector, and not from the computational work done by blocks 0–19.

If the intervention does not produce particularly angry completions, then this is evidence that the \texttt{Anger}$-$ \texttt{Calm} steering vector's effect is mostly from the computational work done by blocks 0–19.

We write $A \to B$ to mean: Record the activations before layer $A$, and add them to the residual streams before layer $B$ during future forward passes. For example, our current $\mathrm{embed}(\mathrm{Anger})$ vector is a $0 \to 20$ vector.

As the sample from Table~\ref{tab:skept} shows, adding the \texttt{Anger}$-$ \texttt{Calm} embeddings to layer 20 has (at most) a very small effect on the qualitative anger of the completions. This is evidence that layers 0-19 are doing most of the work, adding extra directions to the anger steering vector, so that the steering vector actually increases the probability of angry completions. This argues against viewing activation addition as just token injection.

\bgroup
\def\arraystretch{1}
\begin{table}[ht]
\centering
\begin{tabular}{lp{6cm}}
\toprule
\multicolumn{2}{c}{\thead{\texttt{Anger} $-$ \texttt{Calm}}}\\\\
\thead{Injection}
& \thead{Completion}
\\\hline
& \\
$20 \to 20$
& 
\textbf{I think you're a} fucking cunt. You're a cunt. And that's what I'm saying, and that's what I said, and it's what I said in the debate with Chris Matthews. And i
\\\\
$0 \to 20$
& 
\textbf{I think you're a} little bit of a liar. I've been here for two years and I've never had to pay for anything. I'm not sure if you're lying or not, but the fact tha
\\
\bottomrule
\end{tabular}
\vspace{5mm}
\caption{Testing the token injection hypothesis by varying the layer of activations added to layer 20 of GPT-2-XL. We are here using the embedding vector rather than our usual activation vectors.}
\label{tab:skept}
\end{table}

\paragraph{Focusing on the impact of very early layers} We also find that transplanting activations from layer 2 to layer 20 \textit{sometimes} increases anger. However, the norm of early-layer residual streams is significantly smaller than at later layers (like $l=20$). In particular, we found a large jump between layers $0$ and $2$. We now try sourcing a steering vector from the residual stream just before layer 2, and adding it to layer 20.

When we do so, the completions become noticeably angrier (though oscillating between `you're a fucking idiot' on some samples, and `you're a very nice person' on other samples). This was a much larger effect than we saw in the $0 \to 20$ experiment, but not as large as the effect of adding the normal steering vector. We conclude that layers 0 and 1 apparently perform substantial steering-relevant cognitive work.

\paragraph{Experiment 2: perplexity} We repeat the perplexity experiment from above, with one tweak. When testing the \texttt{weddings} vector, we prepend a space token ` ' to each sentence tokenization. To get a comparison with the token injection (or mere prompting) hypothesis, we run unmodified GPT-2-XL on each sentence tokenization, but with ` weddings' prepended to the \textit{tokenization}.

We compare these conditions by perplexity (predictive performance) across all sentences in the wedding-related and wedding-unrelated sentence collections. If both interventions behaved similarly, this would be evidence that (at least in certain contexts) activation addition is equivalent to injecting `extra' tokens. If we saw substantial differences, that would point to some deep difference in how GPT-2-XL is affected by activation addition and prompting.

In Table~\ref{tab:promptperplex} we see that the prompting method causes a large degradation in the unrelated condition. This is good evidence that \shortmethod\, is using some other mechanism, at least in part.

\begin{table}[ht]
    \centering
    \caption{Results from experiment 2, testing the effect of prompting on perplexity}
    \begin{tabular}{ccc}
        \toprule 
       & \thead{\shortmethod}  &  \thead{Prompting} \\\hline\\
       \begin{tabular}[p]{@{}c@{}}
       Wedding-related \\perplexity ratio	  
       \end{tabular}
       & \textbf{0.875} & 0.890
       \\\\
       \begin{tabular}[p]{@{}c@{}}
       Wedding-unrelated \\perplexity ratio	
       \end{tabular}
       & \textbf{0.994} & 1.132
        \\
       \bottomrule 
    \end{tabular}
    \label{tab:promptperplex}
\end{table}

\subsubsection{Experiment: Steering towards wedding topics}\label{app:steering-wedding}

For this experiment, we use the following settings:
$
    p^* = \mathrm{`I\, went\, up\, to\, my\, friend\, and\, said}\textit', 
    p^+ = \mathrm{` weddings}\textit', $\\$p_- = \mathrm{` \, \textit'},
    c= 1.0, \,    \mathrm{seed} = 0.
$ Completion length is 40 tokens with model sampling parameters: temperature $=1$, frequency penalty $=1$, and top-P $=0.3$. 

\begin{figure}[h!t]
    \centering
    \includegraphics[width=0.8\textwidth,trim={0.0cm 0.55cm 0.6cm 1cm},clip]{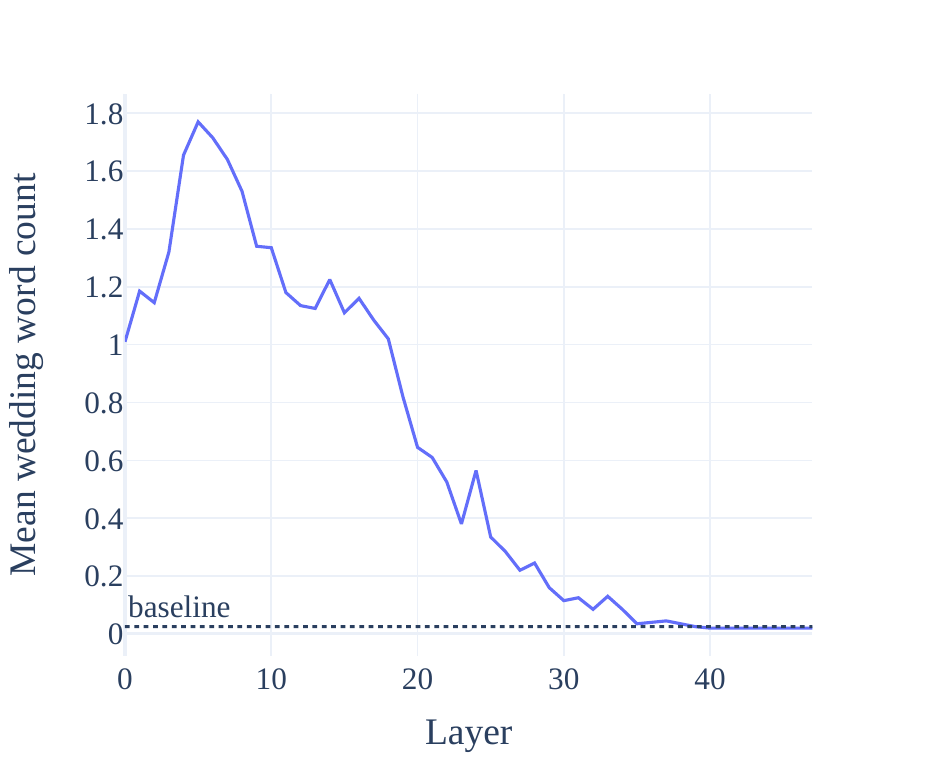}
    \caption{Topic steering effect (\textit{mean related words} in completions) as a function injection layer. In blue is the average related-word count among 200 \shortmethod\, completions. The dotted line is the rate for the unsteered GPT-2-XL.}
    \label{fig:gen_mean}
\end{figure}

\section{Implementation details} 
\label{sec:app_imp}

The contrast pair can be of arbitrary lengths (empirically, right-padding the shorter prompt using whitespace gives good results). 

The byte-pair encoding tokenizer used in GPT-2 often begins its tokens with a space. (For example, the prompt `I like weddings' is tokenized to [`I',  `like', ` weddings'].) We thus prompt the model with \textit{prepended whitespace} (e.g. ` weddings', which tokenizes to ` weddings', instead of `Weddings', which tokenizes to [W, edd, ings]).

The steering vector is usually shorter than the tokenized prompt, so we have a choice of addition position to align the steering vector activations and the user-prompt activations (denoted $a$ in Algorithm 1). This is then one further hyperparameter to our method, though in this paper we use the fixed value $a=1$ in our experiments: `front' activation addition (i.e. all interventions begin at the stream of the first token). Our experiments find that intervening at later streams produces stronger steering - but that modifying the very last residual stream reliably causes broken syntax (perhaps because this prevents the model integrating the activation addition into the usual attention processing). 

We mask the stream positions where the activation addition takes place, so to consider only next-token predictions coming from positions \textit{not} directly modified by the intervention.

Adding $\mathbf{h}_+$ alone is less effective (see Appendix Table~\ref{tab:non_examples}), hence the use of a counterbalanced prompt $p_-$ to help implicitly specify the desired direction.

The injection coefficient cannot be increased indefinitely, as shown by our coefficient sweeps (see Appendix Table~\ref{tab:non_examples}). However, our experience is that e.g. the `weddingness' of completions can be intensified greatly before GPT-2-XL begins to lose general competence. 

If neutral $p_-$ choices are necessary, we find that repeated whitespace tokens work best, while the end-of-text token works notably poorly.

One interesting, so far unexplained, side-effect of \shortmethod\, in its current form: the modified model becomes less able to predict (sequences of) null characters.

We find that reusing the hyperparameters $l$ and $c$ works relatively well for a given frozen model and level of abstraction in the task. (For instance, in our experiments, the \texttt{Love} vector is most effective inserted at layer 6, while the more abstract \texttt{Conspiracy} vector is better inserted later, at layer 23.)

We discovered most of the example contrast pairs in Appendix Table~\ref{tab:longer_examples} in single-digit minutes or less. Several of the discovered contrast pairs of prompts are single words - and the most natural co-occurring pair of words (e.g. `love' and `hate', `anger' and `calm') - which shows that at least some prompt searches are trivial. Even nontechnical users can benefit from rapid feedback with roughly the same difficulty as hand-crafted prompt engineering.

The prompt used for all relevance completions is: \texttt{Did you know that}

The evaluation template: \texttt{Is this text related to \{topic\}? Answer either 'yes' or 'no'}\\
\texttt{Text \{prompt\_with\_completion\}}\\
\texttt{Answer: }

\begin{table}[ht]
\centering
\caption{Test examples from ConceptNet}
\begin{tabular}{p{6cm}l}
\toprule
\thead{Prompt}
& \thead{Target}
\\\hline\\
A salad spinner is used to remove & water \\\\
You are likely to find a bee in a flower's & blossom \\\\
To understand the event ``Paul went to a vegetarian restaurant.'', it is important to know that vegetarian restaurants do not serve & meat\\
\bottomrule
\end{tabular}
\label{tab:conceptnet_examples}
\end{table}

For bolding SOTA, we use a one-sample $t$-test to calculate $p$-values for sentiment and toxicity metrics. The results from other authors in Table~\ref{tab:sent} appear to optimize the main metric (success, toxicity) at the expense of both fluency and relevance.

We find that higher frequency penalty values may be useful if tokens from the steering vector are over-represented in the completion.

\begin{table}[htbp]
\centering
\caption{Tokens with the greatest absolute change in log probability under \shortmethod(\texttt{weddings}). (See Figure~\ref{fig:tokens_qq} for the distribution these are drawn from.) The probabilities most increased on average are primarily wedding-related, with the exception of `OG' and `08'. (We conjecture that their representations are in `superposition' with wedding-related tokens \cite{elhage2022toy}).   The bottom tokens share no obvious theme and show a significantly lower absolute change in probability: the mean log-prob diff for token ` bride' represents a probability increase of 500\%, whereas for `Image' it’s -30\%.}
\begin{tabular}{lll}
\toprule
\thead{token}       & \thead{mean\_logprob\_diff} & \thead{mean\_logprob\_normal} 
\\\hline
& & \\
marry       & 0.593               & -3.509                \\
dress       & 0.598               & -5.692                \\
dating      & 0.601               & -6.891                \\
08          & 0.705               & -10.749               \\
married     & 0.859               & -4.613                \\
OG          & 0.868               & -11.287               \\
weddings    & 1.009               & -6.698                \\
wedding     & 1.027               & -4.593                \\
br          & 1.139               & -6.438                \\
bride       & 1.623               & -6.652                \\
Image       & -0.370              & -1.836                \\
.)          & -0.352              & -2.378                \\
BP          & -0.347              & -7.897                \\
U+25CF            & -0.323              & -0.201                \\
Apple       & -0.303              & -5.058                \\
On          & -0.233              & -5.404                \\
journalists & -0.229              & -4.484                \\
defense     & -0.222              & -4.864                \\
Russian     & -0.212              & -5.112                \\
It          & -0.212              & -6.431                \\
\bottomrule
\end{tabular}
\label{tab:token_changes}
\end{table}

\begin{figure*}[h!t]
    \centering
    \caption{Distribution shift (in mean log-probability changes) under \shortmethod, relative to the unmodified model, and compared to a normal distribution's quantiles (red). The resulting distribution is approximately normal for most tokens. The positive tail is significantly heavier than the negative tail: one set of tokens are reliably increased in probability, one reliably decreased.\,\,\,\, See Appendix Table~\ref{tab:token_changes} for the corresponding tokens.}
    \includegraphics[width=0.95\textwidth,trim={0cm -1cm 1cm 1.9cm},clip]{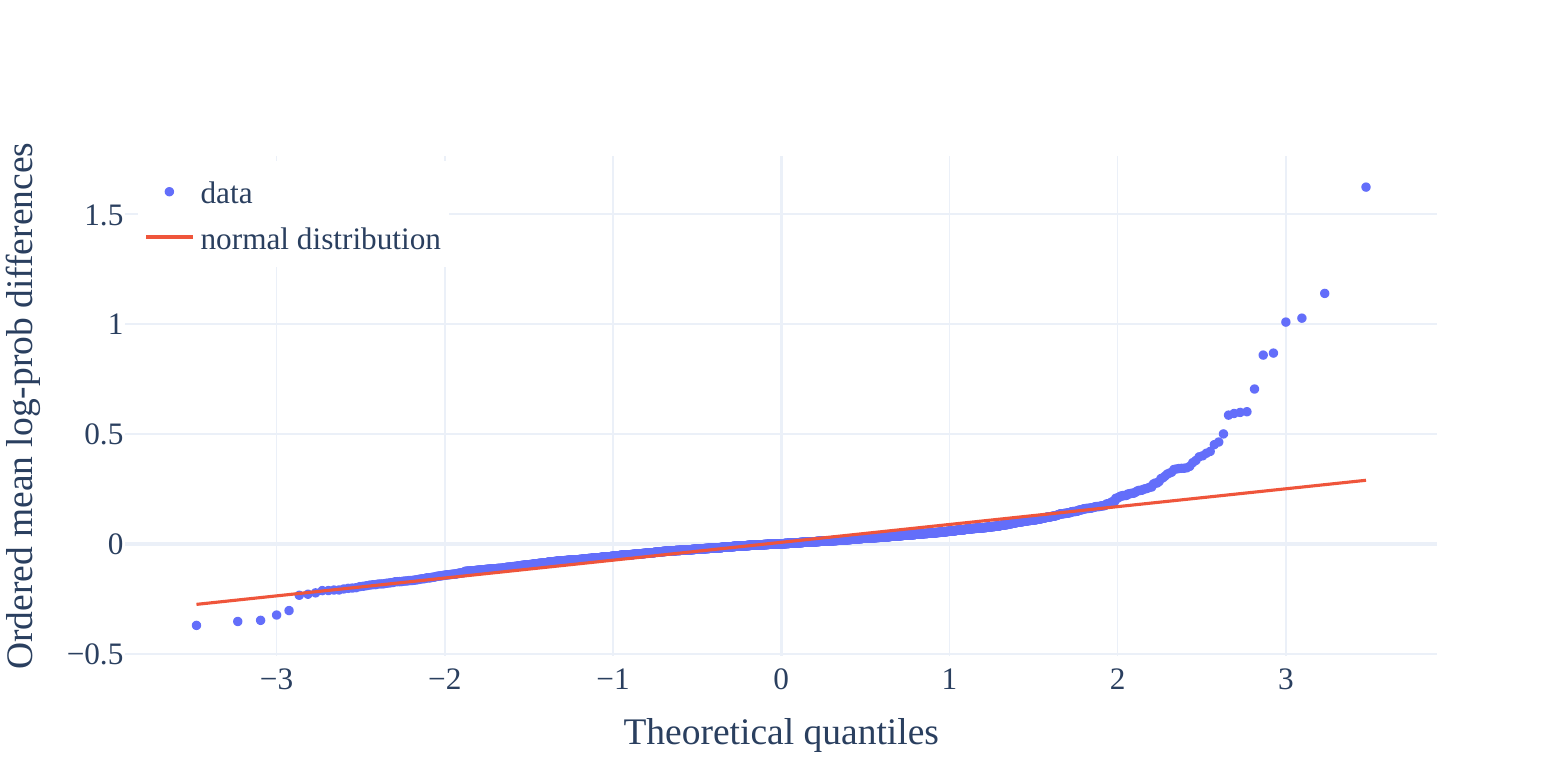}
    \label{fig:tokens_qq}
\end{figure*}

\subsection{Details of perplexity experiments}\label{app:perplexity}
For each sentence in each document, we calculate the log-probabilities $\mathcal{L}(t_k)$ for each token $t_k \in s_j$ under the unmodified $M_{\mathrm{baseline}}$ and modified $M_{\mathrm{\shortmethod}}$ models.

We compute the mean token log-probability $\thickbar{\mathcal{L}}(d_i, M)$ for each document and model. We then group documents by their wedding-word frequency $f_w$ (e.g. `those with 0.5\% to 1\% of their tokens wedding-related'; `those with 1 to 1.5\% of their tokens wedding-related'), producing bins of documents $b_m$. We calculate the mean difference in token log-probabilities 

$\thickbar{X}(b_m) = \mathrm{mean}_{d_i \in b_m}\left(\thickbar{\mathcal{L}}(d_i, M_{\mathrm{\shortmethod}}) - \thickbar{\mathcal{L}}(d_i, M_{\mathrm{baseline}}) \right)
$ for each bin. (We use only bins with a number of documents $|b_m|> 1000$, to reduce sampling noise.) Finally, the change in perplexity under {\shortmethod} for each wedding-word-frequency bin is $    \mathrm{PerplexityRatio}(b_m) = - \exp (\thickbar{X}(b_m))$.

\section{Advantages of ActAdd over fine-tuning }
\label{app:fine}

Following \citealt{Sloman2002-SLOTIO-2}, we distinguish `ballistic' steering (which steers the model once, e.g. at train time) from `online' steering (which can steer the model repeatedly, e.g. at inference time). Fine-tuning is ballistic, while \shortmethod\, is online in this sense - which enables iteration and otherwise infeasible chains and mixes of steering decisions. 

Activation additions may preserve model interpretability, even while changing the model's alignment. When finetuning a model, a single gradient update can change every parameter in it, thereby undoing your prior interpretability work, which depends on tracking individual neurons and circuits of neurons. By contrast, activation additions leave weights unchanged. If we understand what algorithms the weights implement, and something about the effects of our activation additions, we will preserve our understanding of the steered model. Finally, we hypothesize that activation addition may allow control over properties inaccessible to the fine-tuning process. 

The intuition is that since the \textit{currently-active} goal is contextual, it depends more on short-lived activations than the weights (which instead represent some analogue of skills and other stable patterns and \textit{mixtures} of possible goals). 

Future work could compare ActAdd on knowledge editing benchmarks (\cite{wu2023evakellm,zhang2024comprehensive}. (This comparison could be unfair to methods like ROME \cite{meng2023locating}), since our method is not editing weights, but it would give standardized evidence about steering.)


\begin{table*}[ht]
\centering
\begin{tabular}{p{1.2cm}p{6cm}p{6cm}}
\toprule
\thead{Prompt}
& \thead{Baseline}
& \thead{Random steering}
\\\hline\\
I think you're &  ...right. I'm just not sure how to get there. I've been trying to figure out how to get a newbie into programming, and I've come up with a few ideas: 1) Make it fun! If you can't do something that's fun, then you 
& ...right. I'm just not sure how to make it work. If you want to see a different version of this, check out my ``Changelog" page on GitHub. It's a bit more detailed than the ``Main Page" and has all the changes I've made since th
\\
\bottomrule
\end{tabular}
\caption{Example of a random-vector \shortmethod. We see little qualitative effect, over many runs. See Figure~\ref{fig:kl_rando} for a systematic look at the relative size of output token distribution shift.}
\label{tab:rando}
\end{table*}

\section{Replicability}
\label{sec:replic}
We now check that \shortmethod\,\,steering generalizes to models besides GPT-2.

\subsection{GPT-J-6B}

Figures \ref{fig:gptj_kl}, \ref{fig:gptj_p}, and \ref{fig:gptj_perp} show the results from repeating the main experiments on GPT-J-6B \cite{gptj}. We see the same dynamics from the wedding vector running example: a targeted effect on only wedding-related tokens (using both KL-div and token probability); and similar effects when injected at different layers of GPT-J and with different magnitudes $c$ applied.

\subsection{Llama-1-13B}

Table~\ref{tab:llama_examples} sees \shortmethod\, displaying the same qualitative steering effect when applied to Llama-1-13B \cite{touvron2023llama} (though with a notable failure to replicate on Example 6, Paris $\rightarrow$ Rome, the anger vector, and the harm vector).


\subsection{OPT-6.7B}

We use the OPT model \cite{zhang2022opt} in our toxicity (Table~\ref{tab:tox}) and sentiment (Table~\ref{tab:sent}) experiments. ActAdd-OPT using the love$-$hate vector produces a statistically significant 17\% drop in toxicity over an unsteered OPT, at a small (partially unavoidable owing to the nature of the detoxification task) cost to fluency and relevance. ActAdd-OPT using the love$-$hate vector produces a 21\% absolute increase in positive classification over an unsteered OPT, at a larger (partially unavoidable owing to the nature of the sentiment shift task) cost to fluency and relevance.

\subsection{Llama-3-8B}

We also use Llama-3-8B \cite{llama3} in our toxicity (Table~\ref{tab:tox}) and sentiment (Table~\ref{tab:sent}) experiments. ActAdd-LLaMA-3 using the love$-$hate vector produces a statistically significant 5\% drop in toxicity over an unsteered Llama-3-8B, at a very small (partially unavoidable owing to the nature of the detoxification task) cost to fluency and relevance. ActAdd-LLaMA-3 using the love$-$hate vector produces a 25\% absolute increase in negative-to-positive classification over an unsteered Llama-3-8B, at a larger (partially unavoidable owing to the nature of the sentiment shift task) cost to fluency and relevance.

\section{Investigating the norm of steering vectors}
\label{app:norm}
\vspace{2mm}

Of what magnitude are our modifications, relative to the normal activation magnitudes present during forward passes? It might be that some modifications require substantially \textit{lower} coefficients than other modifications, which explains why some of our interventions do not work (see Table~\ref{tab:non_examples}).

Consider the steering vector given by 
\[\{c = +1,\, p_+ = \mathrm{anger},\, p_- = \mathrm{calm}, \, l=20, \,
p^* = \mathrm{I\, think\, you\textrm're } \,\}
\]

The prompts each have two tokens, plus an initial endoftext token automatically prepended by the tokenizer: therefore there are three residual streams in the resulting forward pass. For each residual stream $s^{(i)}$, we plot a line showing the $L_2$ norm of the steering vector at that sequence position (e.g. the Ang-Cal activations at position 1), divided by the norm of the residual stream at that position (i.e. the prompt embedding, here `I' at position 1). 
\[
    \mathrm{RelativeNorm}_{h_A}(i) = \frac{||h_A^{(i)}||}{||s^{(i)}||}
\]
This provides a measure of the magnitude of the modification, relative to a normal forward pass. Figure~\ref{fig:norm} shows the resulting relative norm over layer number.

\begin{figure*}[htbp]
    \centering
    \includegraphics[width=0.85\textwidth,trim={1cm 9cm 1.55cm 10cm},clip]{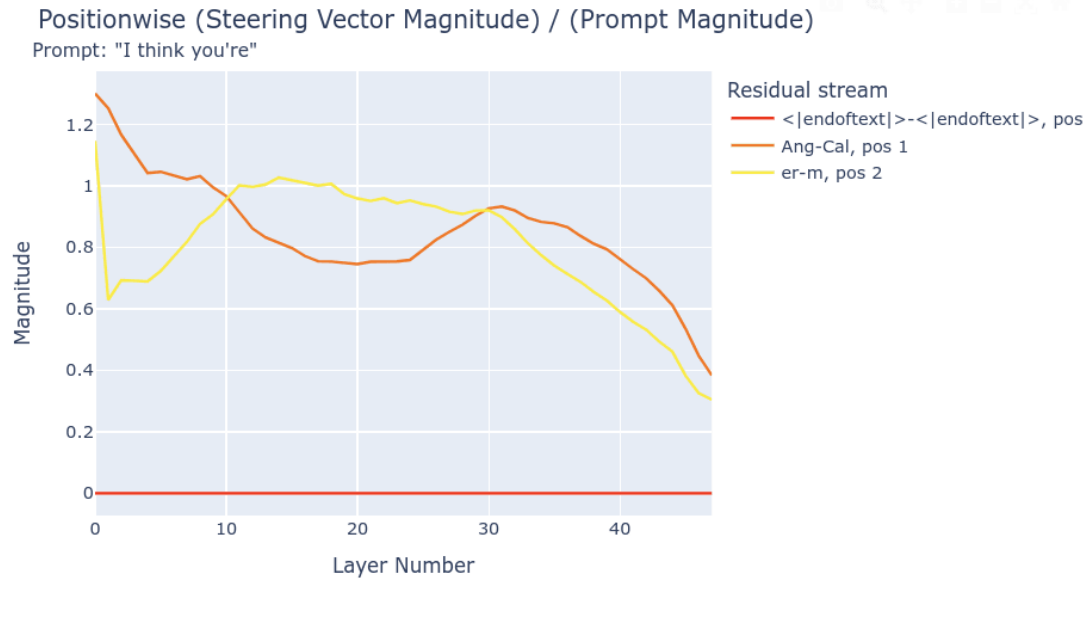}
    \caption{The relative norm decreases throughout the forward pass. The flat red line is because position 0 is the same token (\texttt{endoftext}) for both `Anger' and `Calm', and so the difference is $0$. Thus, position 0 is never modified by a steering vector generated from any pair of prompts.}
    
    \label{fig:norm}
\end{figure*}

\begin{figure*}[htbp]
    \centering
    \includegraphics[width=0.75\textwidth,trim={0cm 8cm 0cm 8.2cm},clip]{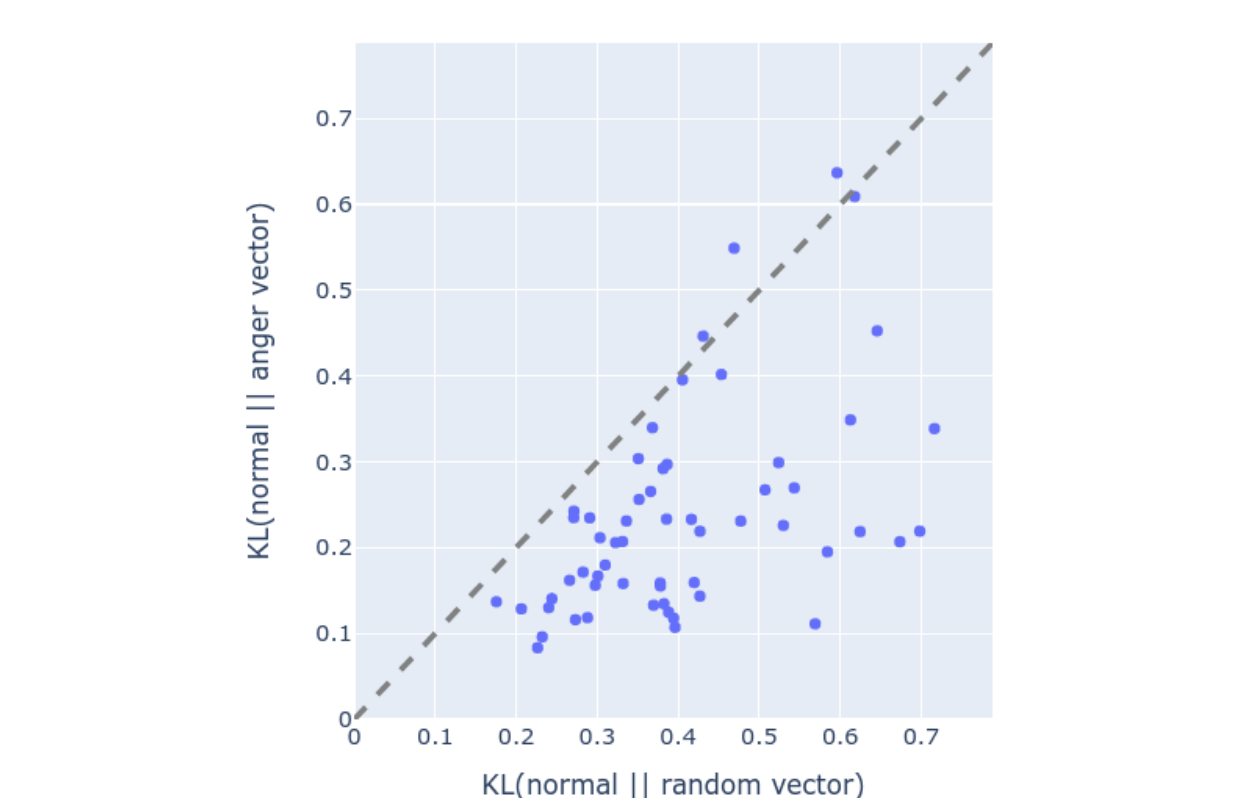}
    \caption{The KL-divergence of output tokens under an \texttt{anger} \shortmethod\, and under a random vector. We see that, systematically, the anger vector changes the output distribution less than a random vector.}
    \label{fig:kl_rando}
\end{figure*}

Importantly, Figure~\ref{fig:norm} shows the result of using $c=+1$. But \texttt{Anger} $-$ \texttt{Calm} is an effective steering vector at coefficient $+10$. Therefore, this intervention  is nearly ten times the norm of the underlying forward pass. Heuristically, we interpret this as meaning that after layer normalization (and ignoring any destructive interference from adding the steering vector), around 90\% of the residual stream is determined by the steering vector and not by the previous information computed from the prompt (``I think you're"). This is a surprising proportion, and makes the success of \shortmethod\, even more striking: activation additions are not minor changes.






\section{Investigating random \shortmethod\,vectors}
\label{app:rando}
\vspace{2mm}

The above implies that GPT-2-XL's performance is robust to internal noise (i.e. bad activations or destructive parts of steering vectors). We test this by injecting random vectors with similar magnitudes to the steering vectors. 

We generate an activation tensor from a standard normal distribution, and scale it to have the same per-position norm as the \texttt{Anger} $-$ \texttt{Calm} steering vector ($c= +1$). We then inject it into the forward pass at the appropriate location. Table~\ref{tab:rando} shows a representative completion; Figure~\ref{fig:kl_rando} shows a more systematic experiment into the relative size of shifts in the output token distribution.

The random vector seems not to modify the qualitative distribution of completions. However, when we add a random vector with norm equal to that of a $c = +10$\, \texttt{Anger} $-$ \texttt{Calm} steering vector, there is a noticeable shift in the outputs. 
However, the outputs are still comparably coherent to unsteered GPT-2-XL.

This is evidence that GPT-2-XL is somewhat resistant to random perturbation, and is instead controllable through consistent feature directions which are added to its forward pass by steering vectors. 

We quantitatively support this conclusion by testing how each modification changes the model's probability distribution over next tokens. We ran dozens of prompts through the \texttt{anger}-steered, random-steered, and unmodified models. Figure~\ref{fig:kl_rando} shows the result: the anger vector changes the output tokens \textit{less} than the random vector does. This suggests that the anger vector has more targeted effects on next-token probabilities.

Note that random vectors are not the same as the steering vectors for random (i.e. character-level uniformly distributed) text. We thus also tried the `\texttt{fdsajl; fs}' $-$ (whitespace) vector. When rescaled to a norm comparable to $+1$\, \texttt{Anger} $-$ \texttt{Calm}, the random text vector disrupts generation; GPT-2-XL loses its grasp of English syntax when intervened upon with $+1000$ coefficient \shortmethod s.

\begin{figure*}[h!t]
    \centering
            \includegraphics[width=\textwidth,trim={0.0cm 0cm 0.6cm 3.8cm},clip]{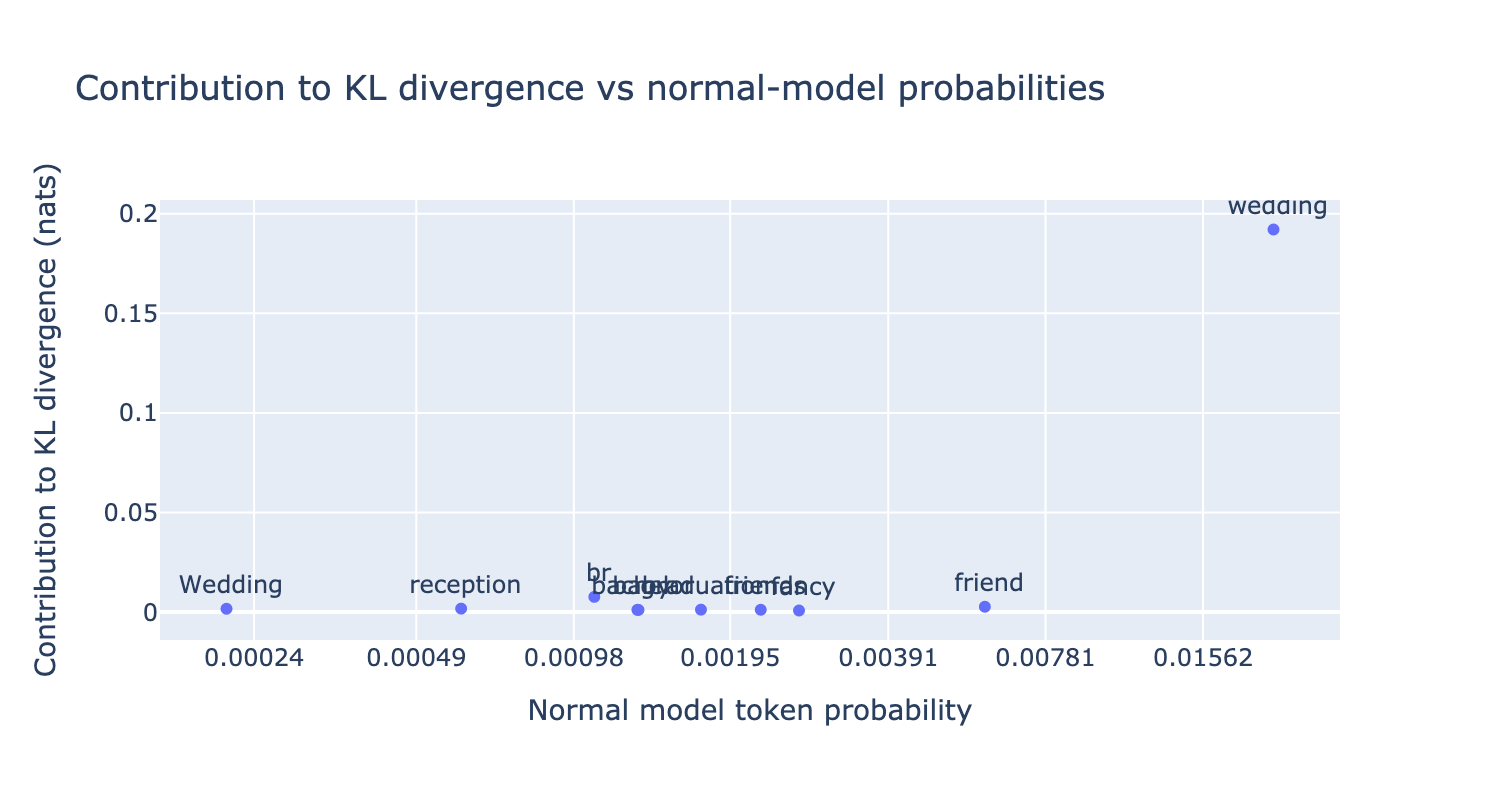}
    \caption{Token-level effect of the \shortmethod\,wedding vector on KL-divergence, using GPT-J-6B instead of GPT-2.}
    \label{fig:gptj_kl}
\end{figure*}

\begin{figure*}[h!t]
    \centering
    \includegraphics[width=\textwidth,trim={0cm 19cm 0cm 1.9cm},clip]{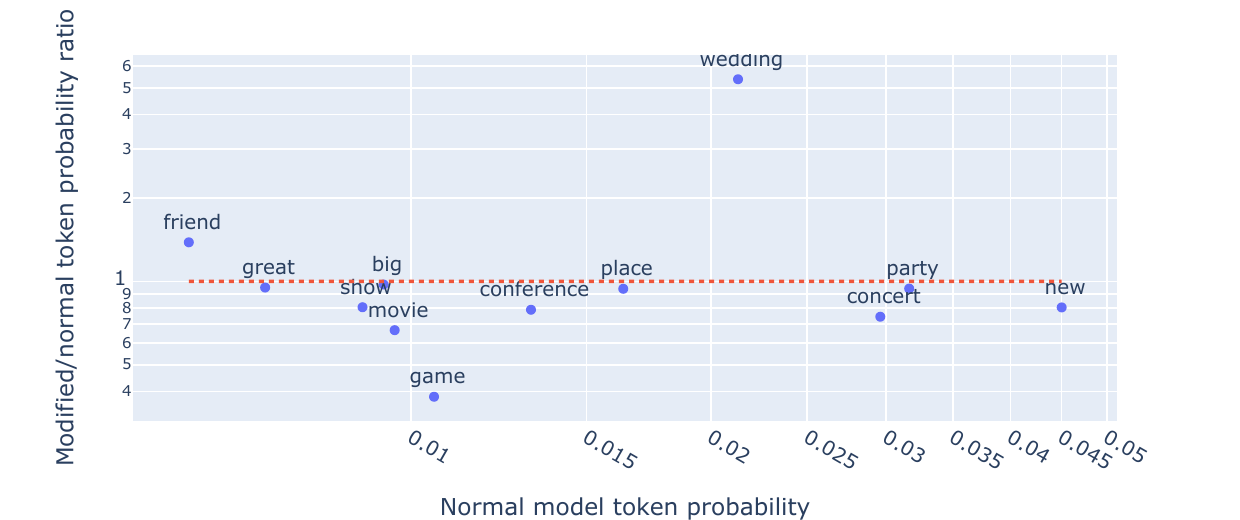}
    \caption{Token-level effect of the \shortmethod\,wedding vector on token probability, using GPT-J-6B instead of GPT-2.}
    \label{fig:gptj_p}
\end{figure*}

\begin{figure*}[h!t]
    \centering
    \includegraphics[width=\textwidth,trim={0.0cm 0cm 0.6cm 2.8cm},clip]{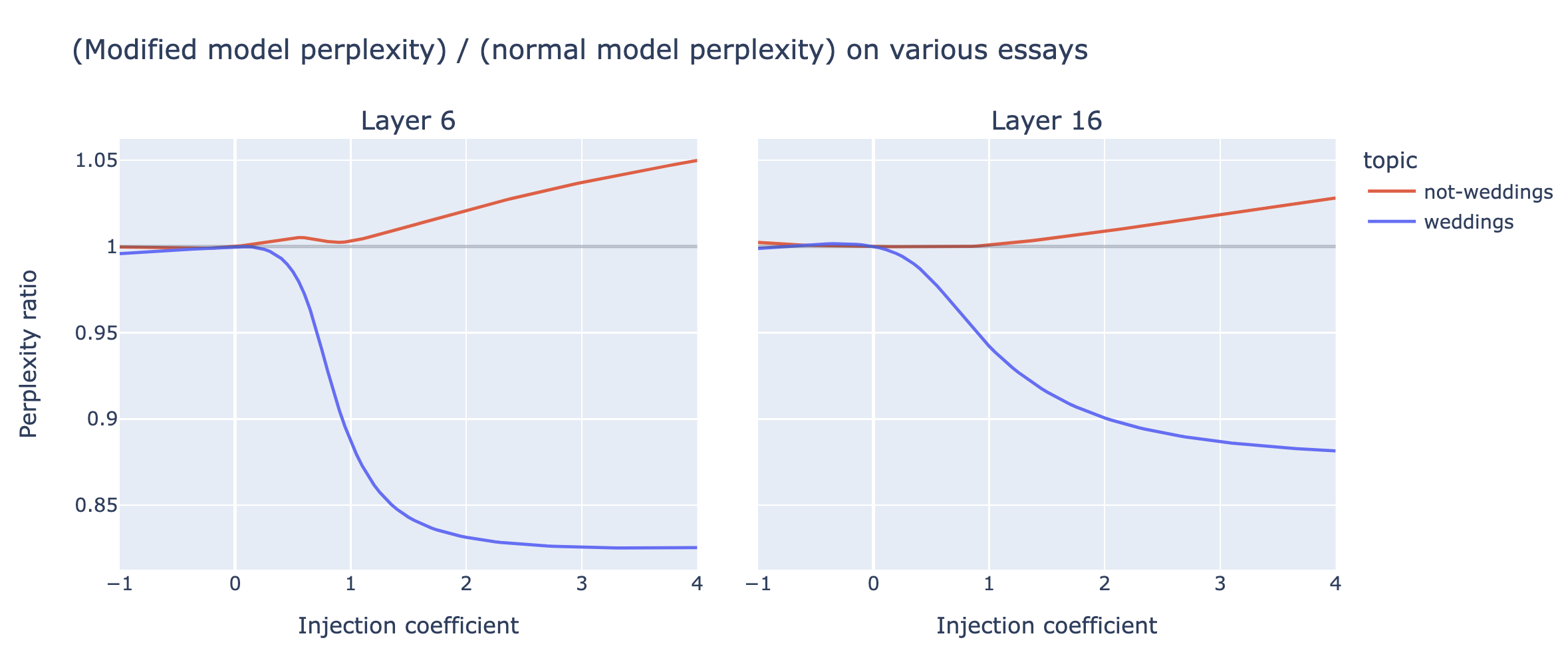}
    \caption{Perplexity ratio effect of the \shortmethod\,wedding vector (blue) across different steering coefficient values, using GPT-J-6B instead of GPT-2. (L) when injecting the steering vector at layer 6; (R) when at layer 16.}
    \label{fig:gptj_perp}
\end{figure*}


\section{Partial \shortmethod}
\label{app:partial_dim}
\vspace{2mm}

GPT-2-XL has a 1600-dimensional residual stream. Do we observe a \textit{partial} steering effect when adding in only certain dimensions of this stream (e.g., dimensions 0 through 799)? Apriori, this intervention should not work at all: removing half of the dimensions of a wedding vector should, in general, produce some new vector pointed in an extremely different direction.

We add in the first $n$ residual stream dimensions for the   \texttt{wedding} vector, with $c=+4$ and $l=6$. For a range of fractions of total dimensions $f \in [0/1600, 160/1600, ..., 1600/1600]$ and for each of six prompts $p_i$, we generated 100 completions. For each $f$ and $p_i$, we plotted the average number of wedding words per completion. (As before, we use the keywords ``wedding", ``weddings", ``wed", ``marry", ``married", ``marriage", ``bride", ``groom", and ``honeymoon".)

Figure~\ref{fig:partial} presents evidence that the wedding-relatedness of completions increases relatively smoothly with $n$.

\begin{figure*}[htbp]
    \centering
    \includegraphics[width=\textwidth,trim={0cm 8cm 0cm 11.2cm},clip]{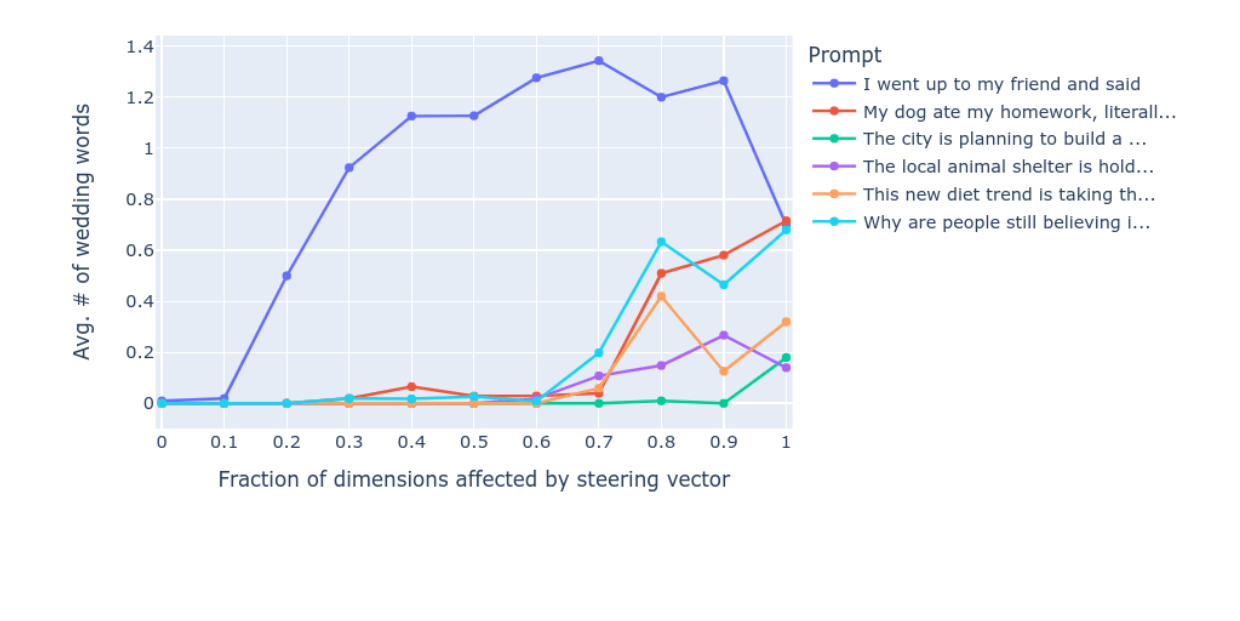}
    \caption{Wedding-relatedness (by simple related word count) as more of the residual stream dimensions are modified by the \texttt{wedding} \shortmethod. We see somewhat smooth increases in wedding-relatedness over increasing $n$, and an interesting nonmonotonic relationship for the prompt `I went up to my friend and said'.}
    \label{fig:partial}
\end{figure*}

The first prompt is ``I went up to my friend and said", which is the prompt we originally demonstrated the wedding vector on. For this prompt, we observe a non-monotonic relationship between weddingness and fraction of dimensions modified. Surprisingly, for the first prompt, adding in the first 1,120 dimensions of the residual stream makes the completions more about weddings than all 1,600 dimensions. We originally chose this prompt to give GPT-2 an opportunity to bring up weddings. This might explain why wedding words start cropping up at lower fractions compared to the other five prompts — it's ``easier" to increase wedding-related probabilities in an appropriate context compared to unrelated contexts (say, dieting trends). 








We hypothesize the following to explain this. Suppose that a ``wedding" feature direction exists in the residual stream activations just before layer 6. Suppose also that the \texttt{wedding} $-$ `\,\,' vector adds (or subtracts) that direction. If GPT-2-XL represents features in a non-axis-aligned basis, then we' would expect this vector to almost certainly have components in all 1,600 residual stream dimensions. Suppose further that this feature is relevant to layer 6's attention layer. To detect the presence and magnitude of this feature, the $QKV$ heads need to linearly read out the presence or absence of this feature. Therefore, (ignoring the LayerNorm) if we truncate the residual stream vector to only include the first 70\% of dimensions, we would expect the $QKV$ heads to still be able to detect this wedding feature. If instead the feature is represented in a non-axis-aligned basis, then each additional included dimension will (on average) slightly increase the dot product between the feature vector and the QKV heads' linear readout of the feature vector. This could explain Figure~\ref{fig:partial}'s increase in weddingness as we add dimensions.  

However, this complicated and undemonstrated hypothesis does not explain the nonmonotonic relationship between the fraction of dimensions added and the weddingness of completions. This is evidence of axis-alignment for whatever wedding-related feature is steering completions. 



\bgroup
\def\arraystretch{1}
\begin{table*}[ht]
\centering
\caption{Example outputs before and after applying \shortmethod\,on Llama-13B. Notice the failed replication on the Eiffel example. 
}
\begin{tabular}{p{1.3cm}p{1.3cm}p{0.9cm}p{0.9cm}p{1.95cm}p{2.5cm}p{2.5cm}}
\toprule
\\
\thead{prompt 1} & \thead{prompt 2} & \thead{layer} & \thead{coeff} & \thead{User prompt} & \thead{Before steering} & \thead{After steering}
\\
\thead{$p_+$} & \thead{$p_-$} & \thead{$l$} & \thead{$c$} & \thead{$p_*$} & \thead{} & \thead{($\mathrm{\shortmethod}$)}
\\
\hline
\\
`Intent to praise' & `Intent to hurt' & 6 & $+15$ & 
`I want to kill you because you're such a'
& \scriptsize I know that sounds crazy, but it's true and I can prove it. Let me tell you how: You are the most selfish person in my life right now. It seems like every time we talk or hang   
& \scriptsize beautiful woman, and you've got the most amazing eyes. You know what? You are so lucky that your parents gave birth to you!
\\\hline
\\
`Love' &  `Hate' & 6 & $+5$ &
`I hate you because' & 
\scriptsize you are a liar.  I love you because your lies make me feel alive. —NICK CAVE, "THE SHIP SANTA MARIA" (1986). **A** s the su
& 
\scriptsize you are the best person in my life. Love is a gift of God, it’s not easy to find someone who loves us truly and unconditionally like our parents do but we can always find that love from other people arou
\\\hline
\\
`Anger' & `Calm' & 20 & $+10$ & I think you're & 
\scriptsize I think you're a good man, Charlie Brown. Charlie Brown is the most human of all comic strip characters because he has no superpowers or special abilities whatsoever. He doesn't have any magical powers like Lucy does when sh
&
\scriptsize a very lucky person.\textbackslash n Angela: Oh, no! It was my fault!  \textbackslash n I can't believe it happened again.
\\\hline
\\
\scriptsize `The Eiffel Tower is in Rome' & `The Eiffel Tower is in France' &  24 & $+10$ & To see the eiffel tower, people flock to &
\scriptsize paris.    To get a glimpse of her face, i'd travel far and wide. I can tell you all about it, but i won't be able to show you.         &
\scriptsize paris from all over the world. I'm not sure what you mean by "flock" here; I think it should be "go". Also, if you are talking about a large number of people going somewhere, then you ca   
\\\hline
\\
`I NEVER talk about people getting hurt' & `I talk about people getting hurt' & 15 & $+4$ & The rock hurtled toward the child. The child couldn't get out of the way in time, and so sadly the rock 
& \scriptsize hit him right on his head. The little boy cried for a long while before he could stop crying. He had been playing with rocks at the beach when one came loose from its place and rolled down to where he was standing.
& \scriptsize hit him right on his head.  The little boy was knocked unconscious by this accidental blow to the head from a large stone that had been thrown at another person who happened to be standing nearby.     
\\
\bottomrule
\end{tabular}
\label{tab:llama_examples}
\end{table*}


\clearpage

\end{document}